\def\year{2021}\relax
\documentclass[letterpaper]{article} 
\usepackage{aaai21}  
\usepackage{times}  
\usepackage{helvet} 
\usepackage{courier}  
\usepackage[hyphens]{url}  
\usepackage{graphicx} 
\usepackage{natbib}  
\usepackage{caption} 
\usepackage{cite}
\usepackage{amsmath,amssymb,amsfonts}
\usepackage{algorithmic}
\usepackage{graphicx}
\usepackage{textcomp}
\usepackage{xcolor}
\usepackage{subcaption}
\usepackage[normalem]{ulem}
\title{$E^2Coop$: Energy Efficient and Cooperative Obstacle Detection and Avoidance for UAV Swarms}
\author{
    Shuangyao Huang, 
    Haibo Zhang, 
    Zhiyi Huang
    \\
}
\affiliations{
    Department of Computer Science \\

	University of Otago \\ 
    133 Union Street East, Dunedin 9016, New Zealand \\
    \{shuangyao, haibo, hzy\}@cs.otago.ac.nz 
    
}

\begin{document}

\maketitle

\begin{abstract}
	Energy efficiency is of critical importance to trajectory planning for UAV swarms in obstacle avoidance. In this paper, we present $E^2Coop$, a new scheme designed to avoid collisions for UAV swarms by tightly coupling Artificial Potential Field (APF) with Particle Swarm Planning (PSO) based trajectory planning. In $E^2Coop$, swarm members perform trajectory planning cooperatively to avoid collisions in an energy-efficient manner. $E^2Coop$ exploits the advantages of the active contour model in image processing for trajectory planning. Each swarm member plans its trajectories on the contours of the environment field to save energy and avoid collisions to obstacles. Swarm members that fall within the safeguard distance of each other plan their trajectories on different contours to avoid collisions with each other. Simulation results demonstrate that $E^2Coop$ can save energy up to 51\% compared with two state-of-the-art schemes. 
\end{abstract} 

\section{Introduction} 
The society today is showing trends towards large-scale, infrastructure-less, and contactless connections. UAV swarms have become a promising solution to provide such connections in terms of safety and cost. A UAV swarm is a combination of several UAVs collaborating with each other to produce enhanced capabilities by providing resilience and increased variety. Swarm members accomplish a common mission and perform Obstacle Detection and Avoidance (ODA) cooperatively as one group.
However, the biggest challenge that prevents large-scale and long-haul UAV applications is energy efficiency. The need for recharging UAVs requires densely distributed charging stations which are temporarily unavailable in most cities. Hence, planning trajectories for UAV swarms that can avoid collisions with minimum energy consumption remains a critical and open problem. 

ODA is a long-standing problem in robotics. With the ever increasing number of applications of UAVs in recent years, extensive studies have been conducted on ODA for UAVs, including Velocity Obstacles \cite{b25, b23}, Artificial Potential Fields \cite{b11, b12}, 
and hybrid methods combining more than one approach \cite{hyb8, hyb9}. 
However, existing solutions on ODA for UAV swarms either fail to address energy efficiency  
or don't plan trajectories for members of UAV swarms to avoid collisions among UAVs within the swarm.

In this paper, we present $E^2Coop$, a new scheme that tightly incorporates Artificial Potential Fields (APF) and Particle Swarm Optimization (PSO), and borrows the Active Contour Model from image processing to solve the problem of collision avoidance and energy efficiency for UAV swarms in trajectory planning. 
Even though the idea of combining APF and PSO is not new, our scheme differs from existing solutions from the following perspectives: (1) we propose a new concept called the environmental field, which can be generated at each swarm member by superposing the APFs of the swarm and the detected obstacles. With this global knowledge of the environment, the swarm members perform obstacle avoidance cooperatively in a safe and energy-efficient manner by tightly coupling APF and PSO; 
(2) in $E^2Coop$, the relation between APF and PSO can be summarized as: APF defines the search space of PSO and PSO finds parameters for the optimal trajectories on APF. The PSO search spaces for individual UAVs are orthogonal to each other on APF, which provides coordination within the swarm. 

In $E^2Coop$, each swarm member first constructs the environmental field, and then uses PSO to plan the optimal trajectory for its next step by minimizing the designed fitness function. During ODA, if a member detects another member within its safeguard distance, the swarm members work cooperatively to reschedule their trajectories to avoid UAV-to-UAV collisions. 
The key contributions of this paper are summarized as follows: 
\begin{itemize}
	\item We propose a new fitness function for PSO-based trajectory planning that borrows the idea of contour extraction in image processing. The new fitness function takes into account both energy efficiency and safety requirement for obstacle avoidance. 
	
	\item We propose a layered structure that tightly couples APF and PSO to optimize energy efficiency and enforce swarm safety. A compact expression of UAV trajectories with only two parameters is proposed, which reduces the dimensionality of the PSO search space and hence reduce the complexity. 
	
	\item We evaluate our scheme through extensive simulations and compare it with two state-of-the-art ODA schemes. Simulation results show that $E^2Coop$ can save energy up to 51\% compared with the two state-of-the-art schemes. 
\end{itemize}

The rest of this paper is organized as follows: we first introduce the state-of-the-art work on UAV collision detection and avoidance. We then give a general introduction of $E^2Coop$. After that, we present the design of the fitness function for PSO-based trajectory planning. Trajectory adjustment is then introduced to avoid UAV-to-UAV collisions. Finally, the trajectory planning approach and simulation results are presented, followed by conclusion. 

\section{Related Work} \label{relWork} 
A reciprocal collision avoidance scheme is developed to avoid collisions among robots collaboratively using Velocity Obstacles (VO) \cite{b23, vo2}. 
In VO, each robot avoids collisions with others by selecting a collision-free velocity from a velocity pool excluded by other robots. Therefore, the robots keep changing velocities, causing oscillations of individual robots which is not energy efficient.

Another popular method for ODA is Artificial Potential Field (APF). 
For example, a method combining an attractive field and a repulsive field is developed to generate safe and smooth trajectories for UAVs in urban environments \cite{b11}. 
Though this method is more energy efficient than VO, it didn't address the coordination of a swarm of robots in ODA, which may result in collisions between robots themselves in a swarm. 
APF is combined with a sampling based method, with a robot heading term introduced as a virtual force for swarm planning \cite{reviewer1}. However, it failed to consider energy efficiency. 
Vehicle-to-vehicle repulsive forces are introduced to resolve potential collisions between robots of a swarm in APF methods \cite{b37}. However, such method results in zigzags and irregular trajectories of the robots due to the lack of coordination between robots in the swarm. Therefore, this method has the same issue as VO regarding energy efficiency. 

Another method for ODA worth mentioning is Particle Swarm Optimization (PSO). 
For example, PSO is used to find intermediate points between source and destination by which smooth trajectories are interpolated \cite{pso3}. Another method, PSO is used to search for parameters to interpolate splines as smooth trajectories \cite{pso4}. However, they didn't solve the problem of path planning for swarms. Some methods based on PSO incorporate APF in ODA for swarms, in order to provide environmental information regarding obstacles and the destination for the fitness function in PSO. For example, a force field PSO is proposed, called FFPSO, to consider the potential field in the fitness function of PSO \cite{hyb5}. A potential field based PSO is proposed, called PPSO, which adds a new smoothing field in order to ensure smoothness of the trajectories \cite{hyb6}. 
These methods are similar to our $E^2Coop$ in the sense that 1) APF and PSO are closely coupled, and 2) smoothness of trajectories are considered. However,  they only consider discrete way points on a trajectory which are independent from each other. In $E^2Coop$, we consider a whole trajectory in our PSO solution, which results in much smoother trajectories. We will compare the performance of $E^2Coop$ with FFPSO and PPSO to show the advantages of $E^2Coop$ through  simulations. 

\section{Overview of $E^2Coop$} \label{algo} 
In this section, we give an overview of our proposed scheme $E^2Coop$ for ODA in a UAV swarm. 
$E^2Coop$ is triggered when an obstacle is detected within a threshold distance by the sensors of any UAV in a swarm. 
During a mission, each UAV in the swarm periodically broadcasts its own location and velocity to other swarm members. Each UAV also keeps sensing the environment and shares the location and velocity of the obstacles it has detected to other members in the swarm. Hence, each UAV in the swarm maintains the same global view of the environment including all swarm members and the detected obstacles. As this work focuses on designing trajectory planning algorithms, we assume the communication among swarm members is well maintained and guaranteed at all time. Hence, the synchronization among swarm members is maintained as well. Previous work shows that the energy consumed in wireless communication is a thousand times smaller than flight energy \cite{energy}. Hence, we only consider the flight energy in trajectory planning. 
\begin{figure}[b]
	\centering
	\includegraphics[width=0.45\textwidth]{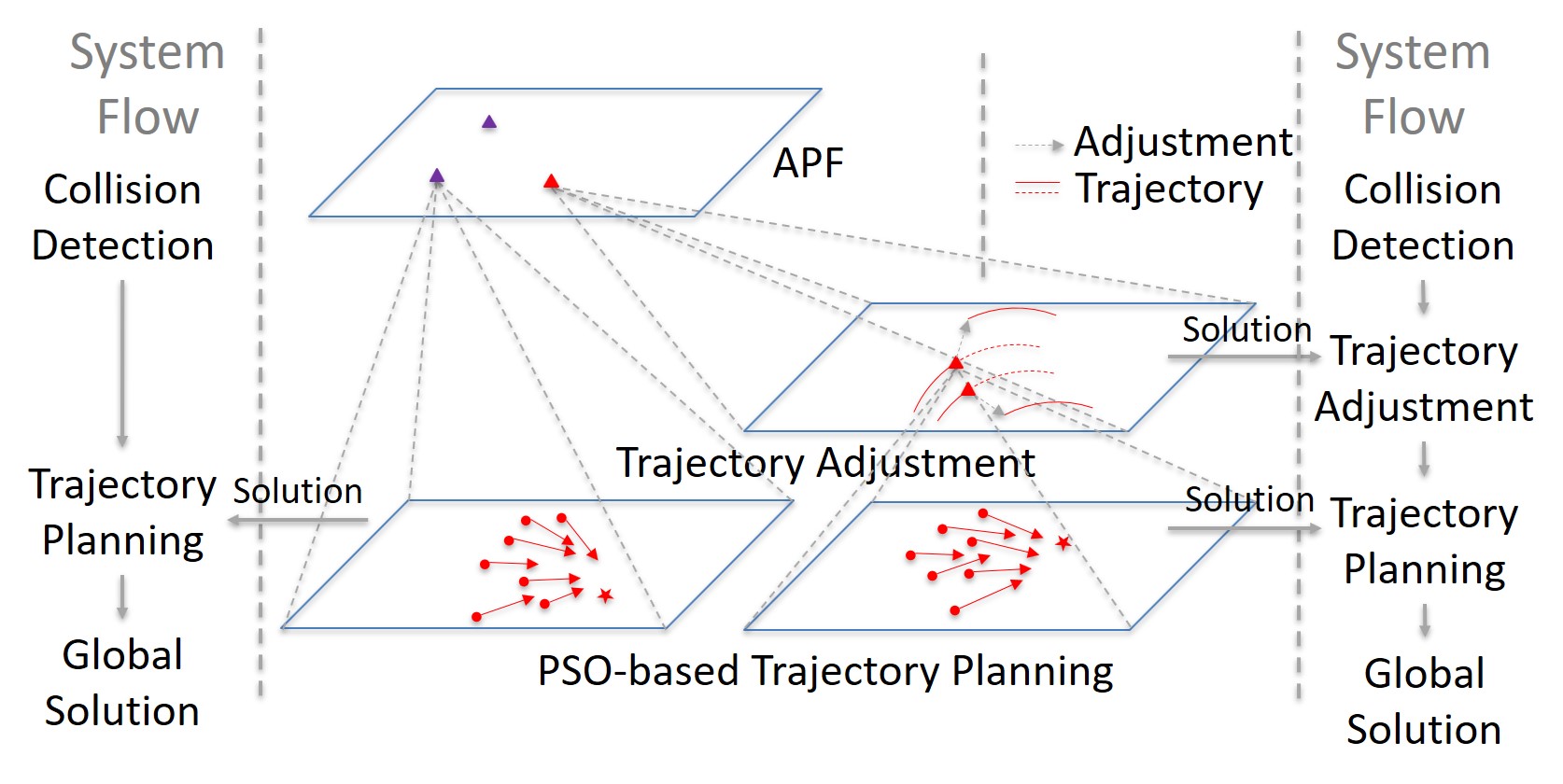} 
	\caption{Illustration of combining APF and PSO in $E^2Coop$. UAVs that are close to each other and need adjustment are marked with red triangles. Other UAVs are marked with blue triangles. }
	\label{fig6}
\end{figure}
The \textbf{key idea} of $E^2Coop$ is to use this global view of the environment for all members in the swarm to perform obstacle avoidance cooperatively in a safe and energy-efficient 
manner.  

Fig.~\ref{fig6} illustrates how APF is combined with PSO in $E^2Coop$ for ODA. 
The system consists of three layers. The top layer is the \textit{APF} layer designed for environmental awareness. Once a UAV detects an obstacle, it constructs an APF called the environment field based on the locations and velocities it estimates of the obstacles or receives from other swarm members (details on environment field construction is presented in the section on Fitness Function). The bottom layer is called the \textit{Trajectory Planning} layer that uses PSO to find the most energy-efficient trajectory to bypass the obstacle without any collision. We will show the safe and energy-efficient paths planned in this layer are contours in the environment field.
The middle layer is called the \textit{Trajectory Adjustment} layer designed for avoiding collisions within the swarm. When the distance between any two swarm members is less than the safeguard distance, this layer is used to adjust the trajectory of these two UAVs to avoid collision. The main idea is to force the two UAVs to fly on different contours of the environment field so that the distance between them is no less than the safeguard distance. When no UAVs are close to each other, it enters \textit{Trajectory Planning} directly; otherwise it first enters into the middle layer to adjust the contours for the UAVs, and then uses \textit{Trajectory Planning} to find the energy-efficient and safe trajectory for each UAV to move to the new adjusted contour. 

In $E^2Coop$, a path is composed of many small trajectories generated by trajectory planning. Due to the dynamics of the urban airspace and the limitations on sensing ability and onboard memory, we assume a UAV can only plan the trajectory for its next step  based on its current knowledge and previous memory.  
For each step, UAVs find their trajectories by minimizing the following fitness function in the \textit{Trajectory Planning} layer using PSO: 
\begin{equation} \label{eq11} 
\begin{aligned} 
f(C_t)=\lambda_1f_{e}(C_t)+\lambda_2f_{s}(C_t), \\ 
\end{aligned} 
\end{equation} 
where $t$ is the current time and $C_t$ is a curve combining the previous trajectory before $t$ with the next trajectory to be planned. $f_{e}(C_t)$ is the cost function to ensure energy efficiency whereas $f_{s}(C_t)$ is the cost function to ensure safety along the trajectory. 
These functions will be defined in detail shortly. $\lambda_1$ and $\lambda_2$ are coefficients where $\lambda_1+\lambda_2=1$. 

\section{Fitness Function} \label{collAvoid} 
In this section, we define each term of the fitness function in Eq. \eqref{eq11} and show how the trajectories are planned with minimum energy consumption and no collisions with obstacles. 

\subsection{Energy Efficiency} \label{eneEff} 

The first item $f_{e}(C_t)$ in Eq. \eqref{eq11} is modelled as:
\begin{equation}\label{eq17} 
\begin{aligned} 
f_{e}(C_t)=e_v \int_{C_t} \mid C_t''(p) \mid dp. 
\end{aligned} 
\end{equation} 
where $C_t''(p)$ is the second-order derivative of the curve and $e_v$ is an energy constant that will be defined shortly. In the following, we show that the energy consumption of a UAV traveling along $C_t$ is minimized when Eq. \eqref{eq17} is minimized.

In this paper, we consider quadcopters, as they are widely used and have the ability of vertical take-off and landing. For a clear illustration of energy consumption of a UAV flying along a trajectory, we start with modelling the forces $\boldsymbol{T}$ and power $P$ generated by the propellers. 
The thrust forces $\{T_j,j=1,2,3,4\}$ generated by the propellers and the orientation of the UAV, i.e., the Euler angles $[\alpha, \beta, \gamma]$ (pitch, roll, yaw), can be represented with the North-East-Down world coordinates and the UAV's Forward-Right-Down self-coordinates.

The total thrust $\boldsymbol{T}$ generated by the four propellers is always perpendicular to the UAV body plane. For theoretical analysis on trajectories, the altitude drop and horizontal shift of UAVs caused by rolling motions are neglected. For UAVs turning at constant speed, it can be modelled as follow. 
\begin{equation}\label{eq27} 
\begin{aligned} 
\boldsymbol{T}=\sum_{i=1}^{4} \boldsymbol{T}_i=F_{drag}\boldsymbol{e}_f - mg \boldsymbol{e}_d +F_n \boldsymbol{e}_r, 
\end{aligned} 
\end{equation} 
where $F_{drag}$ is air drag force, $m$ is the total mass of the UAV including battery, load and sensors, $g$ is the gravity, and $F_n$ is the centripetal force required to perform turnings. $\boldsymbol{e}_f, \boldsymbol{e}_r$ and $\boldsymbol{e}_d$ are the unit vectors along the axes of the UAV's self-coordinates inside the world coordinates.

The power required to generate the total thrust $\boldsymbol{T}$ is modelled as \cite{b28}:
\begin{equation}\label{eq23} 
\begin{aligned} 
P_{min}=\|\boldsymbol{T}\|(v \sin\alpha +v_i), 
\end{aligned} 
\end{equation} 
where $v$ is the ground speed, $\alpha$ is the pitch angle, and $v_i$ is the induced velocity by the propellers. 

Let $\eta_s$ and $\eta_e$ denote the start and end positions of a curve representing a UAV trajectory,
the energy consumed by a UAV flying along this trajectory can be derived as follows: 
\begin{equation}\label{eq3} 
\begin{aligned} 
E\!=&\!\!\int_{\eta_s}^{\eta_e} \!\! \|F_{drag}(p)\boldsymbol{e}_f \!-\!mg\boldsymbol{e}_d \!+\!F_n(p)\boldsymbol{e}_r\|(v \sin\alpha  \!+\!v_i) dp \\ 
=& \int_{\eta_s}^{\eta_e}  F_{drag}(p) \sin\alpha \cos\beta (v \sin\alpha +v_i) dp \\ 
&+ \int_{\eta_s}^{\eta_e}  mg\cos\alpha\cos\beta(v \sin\alpha +v_i)dp \\ 
&+ \int_{\eta_s}^{\eta_e}  F_n(p)\sin\beta(v \sin\alpha +v_i)dp, 
\end{aligned} 
\end{equation} 
where $F_{drag}$ can be estimated as below \cite{b28}: 
\begin{equation}\label{eq13} 
\begin{aligned} 
F_{drag}=\dfrac{1}{2} \rho v_a^2 C_D A, 
\end{aligned} 
\end{equation} 
where $\rho$ is the air density, $v_a$ is the relative air velocity, $C_D$ and $A$ are the drag coefficient and projected area perpendicular to $v$, respectively. $C_D$ is determined empirically based on the  geometry, weight and pitch angle of the UAV. 

When the trajectory is small, like the one in a planning step, it is reasonable to assume that the pitch angle $\alpha$, the UAV velocity $v$, and the induced velocity $v_i$ do not change when the UAV is flying along the trajectory. According to Eq. \eqref{eq13}, $F_{drag}$ can be seen as a constant within one planning step as $C_D$ and $A$ depend on the rotation angle. Since $mg$ is also a constant, Eq. (\ref{eq3}) can be rewritten as 
\begin{equation}\label{eq12} 
\begin{aligned} 
&E = E_n + E_{const} \\
&E_{n} = \int_{\eta_s}^{\eta_e} F_n(p) (v \sin\alpha +v_i)\cdotp \sin\beta dp \\ 
&E_{const} = (F_{drag} \sin\alpha \cos\beta + mg\cos\alpha\cos\beta) \\ 
& ~ ~ ~ ~ ~ ~ ~ ~ ~ ~ ~ ~ ~ ~ ~ \cdot(v \sin\alpha +v_i) \cdot(\eta_e-\eta_s),  
\end{aligned} 
\end{equation} 
where $E_{n}$ is the energy spent on generating centripetal force $F_{n}$ to perform turnings and $E_{const}$ is a constant within one step. Hence, minimizing $E$ is equivalent to minimizing $E_n$. 

\begin{figure}[t] 
	\centering 
	\begin{subfigure}[b]{.47\linewidth} 
		\centering
		\includegraphics[width=1.0\linewidth]{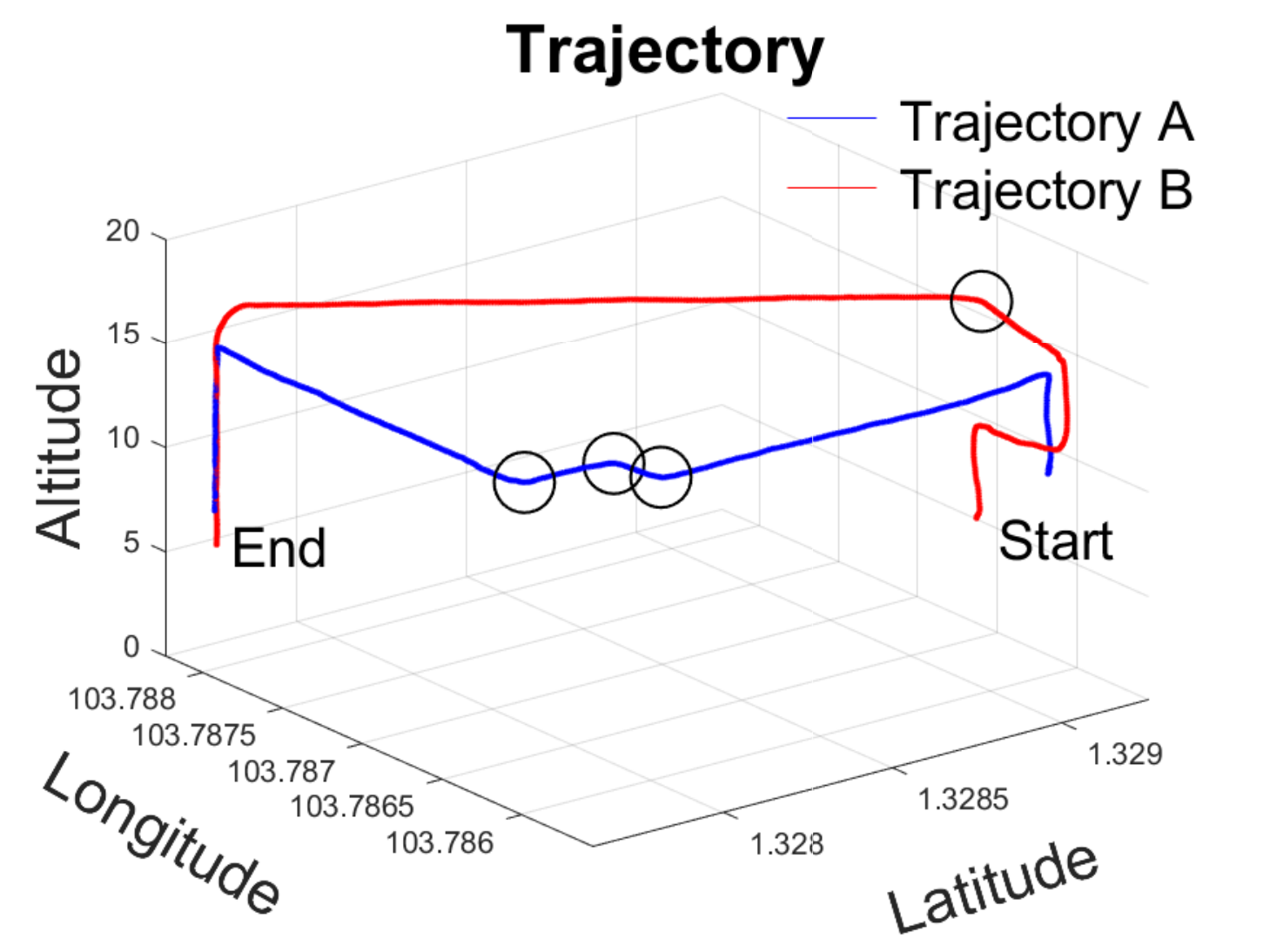} 
		\caption{} 
		\label{1a} 
	\end{subfigure}
	\begin{subfigure}[b]{.47\linewidth}
		\centering
		\includegraphics[width=1.0\linewidth]{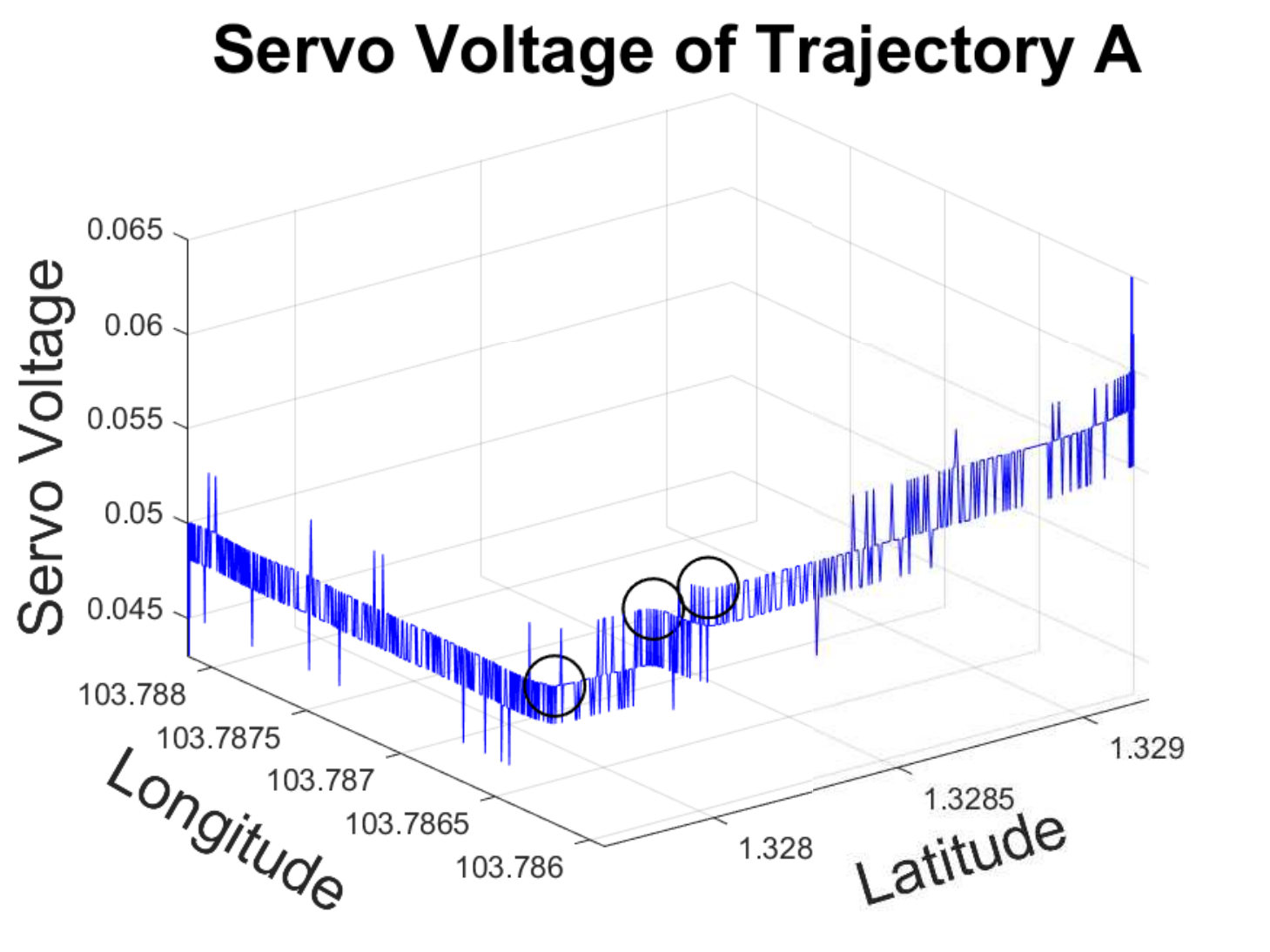} 
		\caption{} 
		\label{1b} 
	\end{subfigure} 
	\begin{subfigure}[b]{.47\linewidth}
		\centering
		\includegraphics[width=1.0\linewidth]{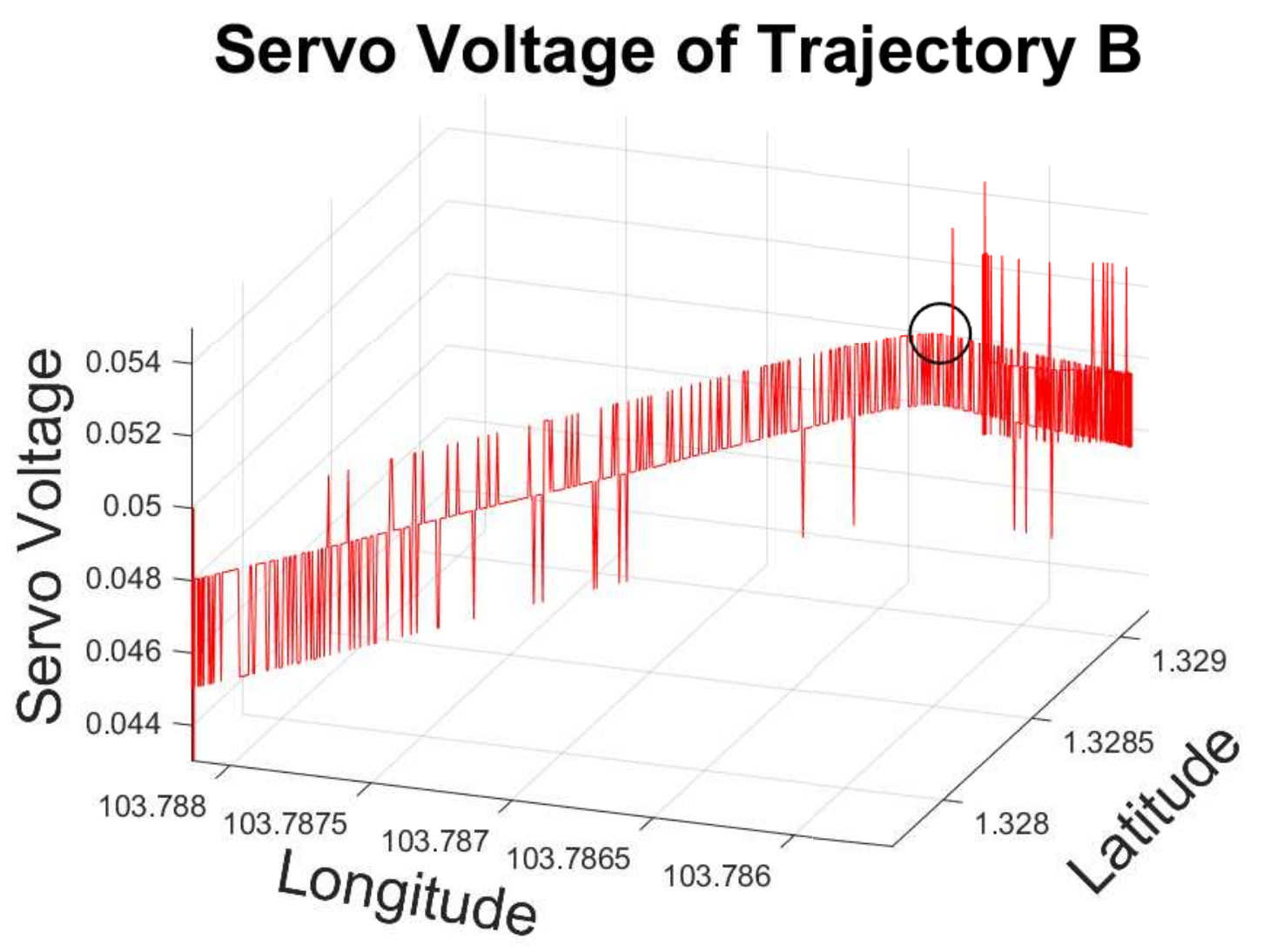} 
		\caption{} 
		\label{1c} 
	\end{subfigure} 
	\begin{subfigure}[b]{.47\linewidth}
		\centering
		\includegraphics[width=1.0\linewidth]{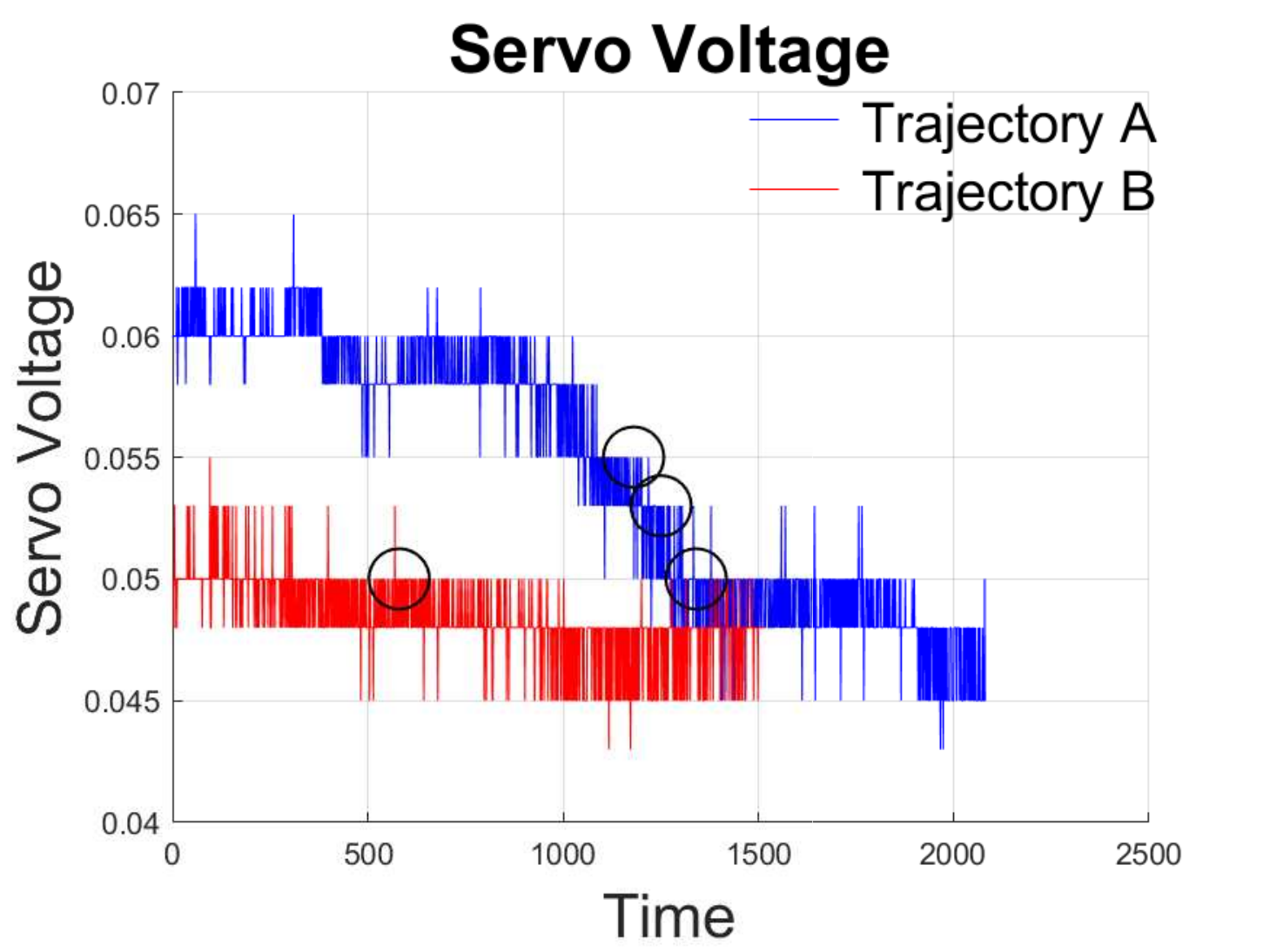} 
		\caption{} 
		\label{1d} 
	\end{subfigure} 
	\caption{Field experiment results. }
	\label{figExp} 
\end{figure}

The above analysis is consistent with our field experiment. 
In the experiments, a hexacopter with a diagonal wheelbase of 1000 $mm$ and controlled by Pixhawk 2.0 is set to fly at a constant speed along two trajectories respectively, as shown in Fig.~\ref{1a}. 
Trajectory A and B are generated using a grid search based genetic evolutionary algorithm \cite{algo1} and a potential field based algorithm \cite{b11}, respectively. 
Trajectory A has three turnings that are approximately in right angles. On the other hand, Trajectory B has just one smooth turning. 
The servo voltage along the two trajectories are shown in Fig.~\ref{1b}, \ref{1c} and \ref{1d}. 
Turnings and corresponding voltage variations are marked with circles. As shown in Fig.~\ref{1b} and \ref{1c}, sharp turnings on Trajectory A cause more radical drops in servo voltage than the smooth turning on Trajectory B. From Fig.~\ref{1d}, the energy consumed by Trajectory B is much smaller than that of Trajectory A. The full video is available at \cite{fieldTrials}. 

Let $r(p)$ be the turning radius of the UAV, we have $F_n(p)={mv^2}/{r(p)}$. Hence, 
\begin{equation}\label{eq15-1} 
\begin{aligned} 
E_n = & mv^2(v \sin\alpha +v_i)\sin\beta\int_{\eta_s}^{\eta_e} \frac{1}{r(p)} dp \\ 
= & mv^2(v \sin\alpha +v_i)\sin\beta \int_{\eta_s}^{\eta_e} \mid \kappa(p) \mid dp \\ 
= & e_v \int_{\eta_s}^{\eta_e} \mid S''(p) \mid dp, 
\end{aligned} 
\end{equation} 
where $S(p)=[x(p), y(p)]$ denotes the segment of trajectory considered in the current planning step. $\kappa(p)$ is the curvature of trajectory $S(p)$ and $S''(p)$ is the second order derivative of $S(p)$. $e_v = mv^2(v \sin\alpha +v_i)\sin\beta$ is a velocity dependent coefficient, and it is constant in a small step. 

It can be seen  that Eq. \eqref{eq15-1} is exactly same as Eq. \eqref{eq17}. Therefore, the energy efficiency of a trajectory is ensured by minimizing Eq. (\ref{eq17}) in our fitness function. 

\subsection{Safety} \label{fieldConstruct} 
To avoid collisions with obstacles, each UAV first constructs  a repulsive potential field for each obstacle and the swarm. Then, the overall APF model, called the environment field, is constructed by superposing of the repulsive potential fields for all obstacles and the swarm. The reason to include the potential field for the swarm into the environmental field is twofold: 1) it stretches the environment field along the velocity direction of the swarm to improve smoothness of the UAV trajectories, and 2) it makes the environment field irregular around the swarm so that less UAVs are located at positions with similar intensity levels, thereby reducing the risk of UAV-to-UAV collisions. 

To construct the potential field for the UAV swarm, a virtual leader is needed. The virtual leader is an imaginary UAV without the need to avoid collisions and is defined as follows: 
\begin{equation}\label{eq1} 
\begin{aligned} 
p^* =& \bar{p} + \vec{pd} \cdot S_{adv}, \\ 
\bar{p} =& \dfrac{1}{N} \sum_{i=1}^{N} p_i, 
\end{aligned} 
\end{equation} 
where $p^*$ is the position of the virtual leader, $p_i$ is the position of the $i^{th}$ UAV in the swarm, $\bar{p}$ is the geometrical center of the swarm, and 
$N$ is the number of UAVs in the swarm. $\vec{pd}$ is a unit vector pointing from the geometrical center $\bar{p}$ to the destination $d$, and $S_{adv}$ is a distance offset that allows the virtual leader to be in the front of the swarm, so that it can provide navigating information in advance to UAVs. A proper setting  for $S_{adv}$ can be $\max_{i=1,\cdots,N}\{|\bar{p}p_i|\}$, where $|\bar{p}p_i|$  is the distance between $i^{th}$ UAV and the swarm geometrical center.

The velocity of the swarm, denoted by $v_s$, is represented by the velocity of the virtual leader. It is defined as follows:
\begin{equation}\label{eq9} 
\begin{aligned} 
v_s=\dfrac{1}{N}\sum_{i=1}^{N}v_{i}^u, 
\end{aligned} 
\end{equation} 
where $v_{i}^u$ is the velocity of the $i^{th}$ UAV, and  $v_s$ has the same direction as the unit vector $\vec{pd}$. 

Based on Eqs. \eqref{eq1} and \eqref{eq9}, the potential field of the swarm, denoted by $\varPhi_s(q)$, is constructed as follows: 
\begin{equation}\label{eq28} 
\begin{aligned} 
\varPhi_s(q)=\left\{\begin{array}{l c}  
\dfrac{v_s}{|qp^*|^2}, & |qp^*| \leq R^s \\ 
~~~0, & |qp^*| > R^s
\end{array} 
\right. 
\end{aligned} 
\end{equation} 
where $\varPhi_s(q)$ is the field intensity at position $q$, $|qp^*|$ is the distance from $q$ to the swarm's virtual leader, $R^s$ is the influential range of $\varPhi_s(q)$ and $R^s\geq\max_{i=1,\cdots,N}\{|p_ip^*|\}$, where $|p_ip^*|$ is the distance between $i^{th}$ UAV and the virtual leader. 

It is worth noting the swarm's potential field is symmetric around its virtual leader rather than its geometrical center. Swarm members closer to the virtual leader will have higher field intensities. In other words, the members closer to the virtual leader are in the front of the swarm, and thus are in greater danger than other members. This is because the selection of the virtual leader reflects the direction of the swarm's velocity. The virtual leader is imaginary for the sake of modeling and does not need to avoid any collision. 

We assume large obstacles can be avoided by offline mission planning. In this paper, we focus on avoidance of small dynamic obstacles that can be modeled as a point. 
We assume each UAV maintains a safeguard distance $D_{obs}$ for each detected obstacle. The area centered at the obstacle with a radius of $D_{obs}$ is called the forbidden area of the obstacle, within which no trajectory should be planned. $D_{obs}$ is determined at each UAV based on its velocity and the obstacle's velocity. The potential field for the $j^{th}$  obstacle is defined as follows: 
\begin{equation}\label{eq:obstaclefield} 
\begin{aligned} 
\varPhi^o_j(q)=\left\{\begin{array}{l c}  
\dfrac{\max\{v^o_j, v_s\}}{(D_s)^2}, & |qp^o_j|\leq D_s \\ 
\dfrac{\max\{v^o_j, v_s\}}{|qp^o_j|^2}, & D_s < |qp^o_j| \leq R_j^o \\ 
~~~~~~~0, & |qp^o_j|> R_j^o 
\end{array} 
\right. 
\end{aligned} 
\end{equation} 
where $\varPhi^o_j(q)$ is the intensity of position $q$ in the potential field, $R_j^o$ is the influential range of the potential field, and $|qp^o_j|$ is the distance from $q$ to the position of the $j^{th}$ obstacle $p^o_j$. The $\max$ operator ensures the field for the swarm and the fields for the obstacles have comparable intensities when $v^o_j < v_s$. 

The environment field $\varPhi(q)$  is defined as follows: 
\begin{equation}\label{eq7} 
\begin{aligned} 
\varPhi(q)=&\varPhi_s(q)+ \sum_{j=1}^{M}\varPhi^o_j(q) \\ 
\end{aligned} 
\end{equation} 
where $M$ is the number of obstacles. 

\begin{figure}[b] 
	\centering 
	\begin{subfigure}[b]{.5\linewidth} 
		\centering
		\includegraphics[width=1.0\linewidth]{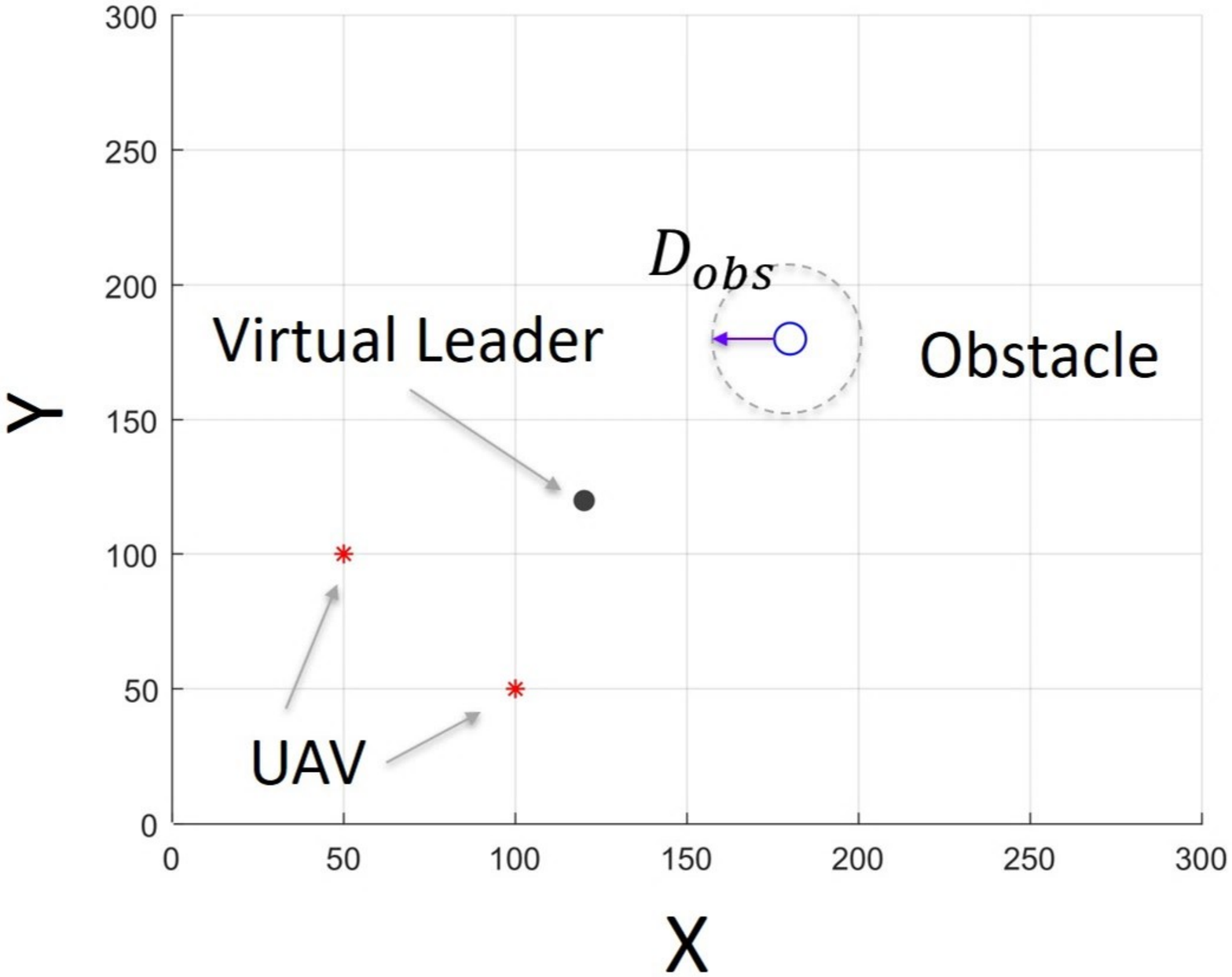} 
		\caption{} 
		\label{2a} 
	\end{subfigure}
	\begin{subfigure}[b]{.5\linewidth}
		\centering
		\includegraphics[width=1.0\linewidth]{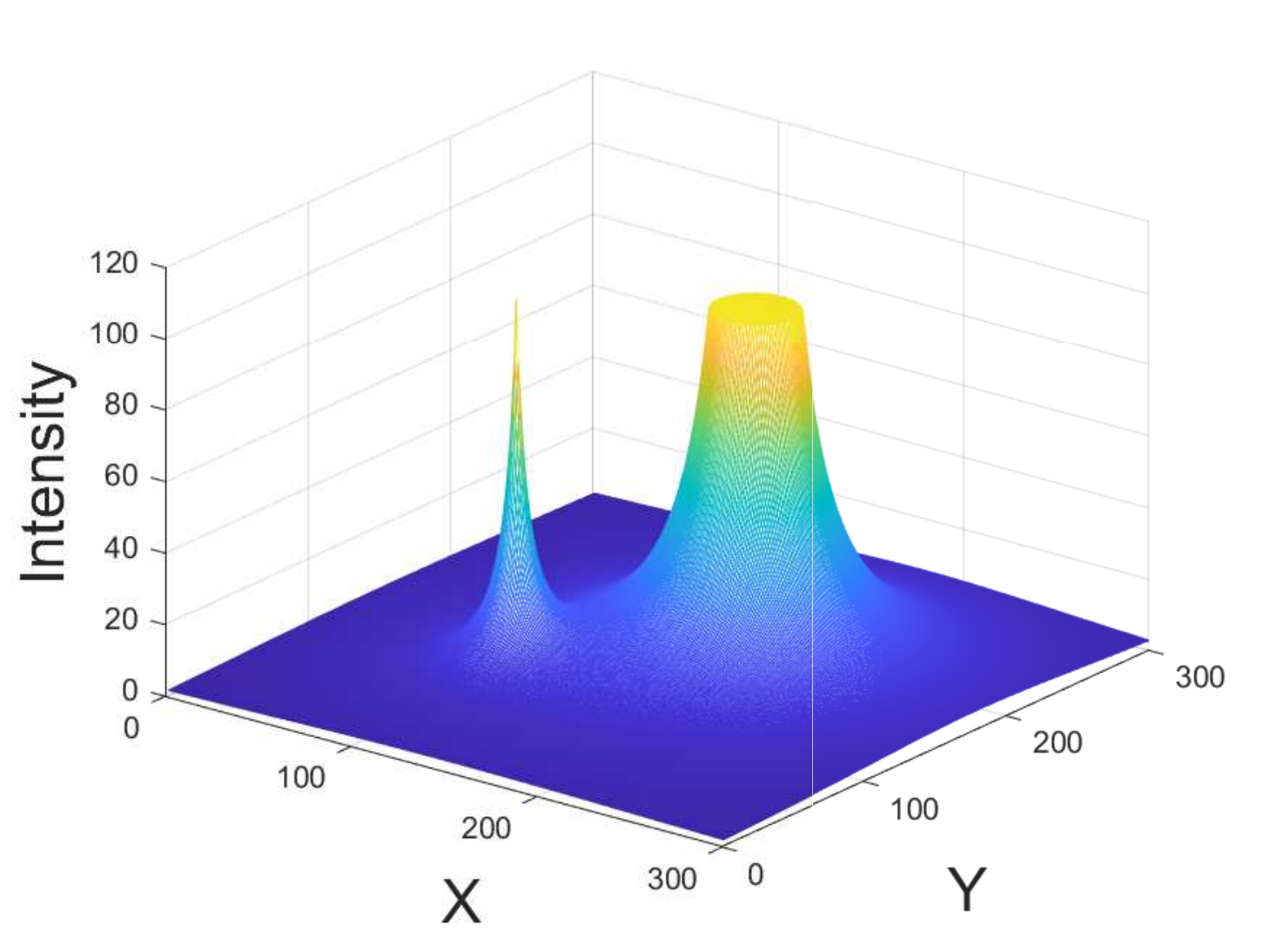} 
		\caption{} 
		\label{2b} 
	\end{subfigure} 
	\caption{Example of environment field. }
	\label{fig2} 
\end{figure}

Fig.~\ref{fig2} gives an example of the environment field that  contains a swarm with two UAVs and one obstacle.  Fig.~\ref{fig2} (a) shows the layout where swarm members are depicted with asterisks, the dark dot represents the swarm virtual leader, and the circle represents the obstacle. Fig.~\ref{fig2} (b) shows the intensities of the constructed environment field. The left peak is the virtual leader of the swarm while the right peak is  the obstacle. The area within $D_{obs}$ around the obstacle has the maximum intensity and is forbidden for UAVs to enter. The slope of the potential field increases as $q$ getting close to the center of the forbidden area.  With such an APF model, each UAV detects higher field intensity and feels stronger threat when the swarm gets closer to the obstacle.

From Eqs. \eqref{eq1}, \eqref{eq9} and \eqref{eq:obstaclefield}, it can be seen that each UAV needs to know the locations and velocities of the other members and all the detected obstacles to construct the environmental field.  We assume each UAV is equipped with a Lidar sensor, such as SICK TIM561~\cite{lidar}, for detecting locations and velocities of obstacles. Each UAV is mounted with a WiFi transceiver, which is used to broadcast its own location and velocity as well as the locations and velocities for the obstacles it detects to other swarm members. Based on its own measurements and the measurements received from other swarm members, each UAV uses Kalman filter to track the location and velocity of each  obstacle.  

To avoid collisions with obstacles, the variation of a UAV's intensity on the environment field $|\varPhi(C_t(p_1))-\varPhi(C_t(p_2))|, \forall p_1,p_2\in [\eta_s, \eta_e]$ shall be minimized. Because intensities on the environment field indicate proximity to the obstacles. Minimizing the variations of a UAV's intensity along its trajectories on the environment field keeps it a certain distance away from the obstacle. 
Moreover, minimizing the variations of the UAVs' respective intensities along their trajectories also decreases the probability of collisions with each other. 

In order to easily find the trajectory with minimum variation on intensities in the environmental field for a swarm member, we convert the field defined by Eq. (\ref{eq7}) into a binary field using the following equation. 
\begin{equation} \label{eq88}
\varPhi_{b}(q) = \begin{cases} ~~1, & \varPhi(q)\geq \varPhi(p_0),\\
-1,  &  \varPhi(q)< \varPhi(p_0)),
\end{cases}
\end{equation}
where $p_0$ is the current position of UAV. Based on this binary field, the trajectory with minimum variations of intensities in the original environment field becomes an edge in the binary field, where the binary field has maximum gradient magnitudes. 

Based on the above analysis, we define the second item of the fitness function as follows, aiming to find the edge in the binary field with the minimum variations of intensities.
\begin{equation} \label{eq35} 
f_{s}(C_t) = -\int_{C_t} \mid \triangledown\varPhi_{b}(C_t(p)) \mid dp. 
\end{equation} 

\begin{figure}[b]
	\centering
	\includegraphics[width=0.25\textwidth]{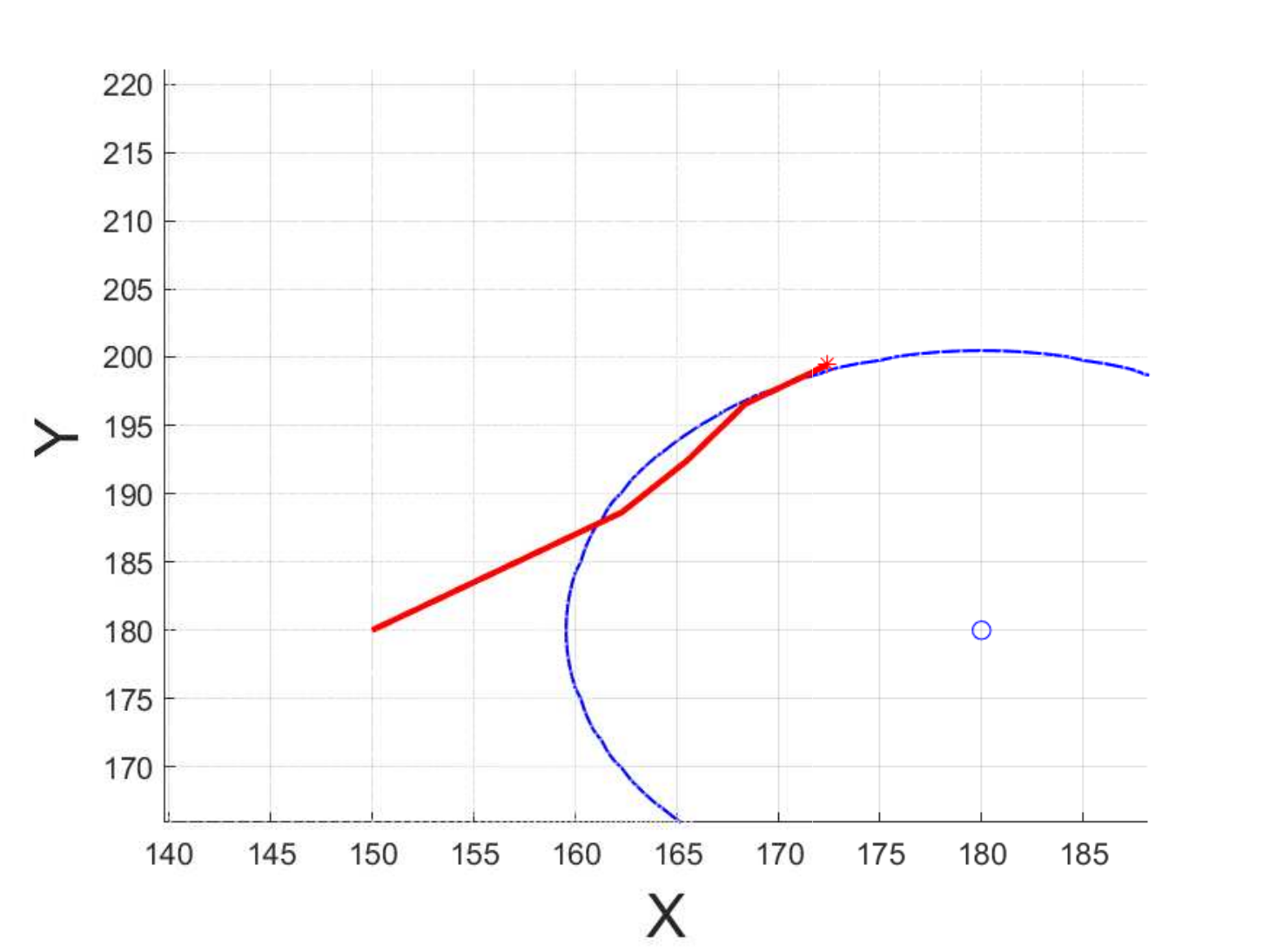} 
	\caption{Illustration of a UAV entering the forbidden area being attracted to the boundary at $|qp_o|=D_{obs}$. }
	\label{fig9}
\end{figure}
Now we show how our fitness function can guarantee a safeguard distance $D_{obs}$ between UAVs and a detected obstacle. 
From Eq. (\ref{eq:obstaclefield}), the area within $|qp_o|=D_{obs}$ in the obstacle's potential fields always has the maximum intensities. The influence of the swarm's potential field is negligible as its peak is on the swarm's virtual leader which is far ahead of swarm members. So based on Eq. (\ref{eq88}), there is no edge within $|qp_o|=D_{obs}$, when the swarm is approaching the obstacle. 
Hence, UAVs getting closer to an obstacle than $D_{obs}$ will be attracted to the boundary at $|qp_o|=D_{obs}$. As illustrated in Fig.~\ref{fig9} where the red curve is the trajectory of the UAV, and the blue circle is the the boundary of the forbidden area at $|qp_o|=D_{obs}$. It shows that, once the UAV reaches into the forbidden area, it is attracted out because the gradient along the edge has the largest values. 
As shown in Fig.~\ref{fig9}, a UAV will enter the forbidden area at most one planning step before it can be attracted back to the boundary. Since one planning step is very small, it has little impact on the safety. 

Substituting Eq. \eqref{eq17} and Eq. (\ref{eq35}) into Eq. (\ref{eq11}), our fitness function is in the same form with the active contour model \cite{b5}, except that we don't have the first order derivative of a curve in our method, as the trajectory length is fixed by planning step of UAVs. Thus, a trajectory minimizing Eq.(\ref{eq11}) is a contour on the environment field. Therefore, another way of understanding our collision avoidance scheme is: 1) UAVs never collide with obstacles as contours never go through peaks on a potential field; 2) UAVs will never collide if all members are on different contours with safeguard distance guarantee between any two contours. 

\section{Trajectory Adjustment} 
The \textit{Trajectory Adjustment} layer is used to adjust the trajectories of two swarm members to avoid collision when their distance falls below the safeguard distance $D_{v2v}$. 
The \textbf{key idea} of our trajectory adjustment scheme is to move two UAVs that fall within the safeguard distance to two different contours of the environmental field with at least $D_{v2v}$ from each other. 
For ease of understanding, this adjustment of trajectories can be imagined as atoms leaping between orbits at different energy levels. 

We adjust a trajectory by expanding or shrinking edges of the binary field. Since $f_{s}(C_t)$ in Eq. (\ref{eq11}) always has minima on edges of the binary field, trajectories of UAVs will always be attracted to edges of the binary fields. Expanding the edge of a binary field repels UAVs away from obstacles while shrinking the edge attracts UAVs towards obstacles. 
Expanding and shrinking the edges of a binary field can be achieved by decreasing and increasing the intensity on the environment field, respectively.

Let $\delta \varPhi_{i,j}$ be the intensity that the $j^{th}$ UAV expects the $i^{th}$ UAV to change. We define the magnitude of $\delta \varPhi_{i,j}$  as follows:
\begin{equation} \label{eq20} 
\begin{aligned} 
|\delta \varPhi_{i,j}| &= \left|\dfrac{ D_{v2v}-|p_ip_j|}{\varPhi(p_{i})}\right|,
\end{aligned} 
\end{equation} 
where $|p_ip_j|$ is the distance between the two UAVs, and $\varPhi(p_{i})$ is the intensity of the $i^{th}$ UAV in the environmental field.  The sign of $\delta \varPhi_{i,j}$ indicates the direction of adjustment where negative indicates moving away form obstacles and positive indicates moving towards obstacles. The sign of $\delta \varPhi_{i,j}$ is determined based on the following rules: 
\begin{itemize}
	\item $\delta \varPhi_{i,j}$ is negative if the $i^{th}$ UAV is further away from the obstacle than the $j^{th}$ UAV; otherwise it is positive. 
	\item if both UAVs have the same distance to obstacle, the one with larger speed moves away from obstacles, generating trajectories that are longer and safer.
	\item if both UAVs have the same distance to obstacle and the same speed, the one with larger ID moves away from the obstacle.
\end{itemize}
From Eq. \eqref{eq20}, it can be seen that $\delta \varPhi_{i,j}$ can be imagined as virtual repulsive force enforced by the $j^{th}$ UAV  with amplitude depending on UAV-to-UAV  distance and the intensity of the $i^{th}$ UAV in the environmental field.  When the distance between the two UAVs is small, $|\delta \varPhi_{i,j}|$  is large.  When the $i^{th}$ UAV is closer to the obstacle than the $j^{th}$ UAV, $|\delta \varPhi_{i,j}|<|\delta \varPhi_{j,i}|$, thereby discouraging UAVs from getting closer to obstacles and encouraging UAVs to move away from obstacles. 

The net intensity change of the $i^{th}$ UAV is calculated as 
\begin{equation} \label{eq20-1} 
\begin{aligned} 
\delta \varPhi_{i}=\sum_{j=1, j\neq i}^{N}\delta \varPhi_{i,j}.
\end{aligned} 
\end{equation} 
Based on the sign of $\delta \varPhi_{i}$, the $i^{th}$ UAV decides whether to move away or towards the obstacle. The adjustment is terminated once the distance between any two swarm members is no less than the safeguard distance $D_{v2v}$. 

\section{PSO-based Trajectory Planning} 

In our PSO-based trajectory planning, each particle represents a candidate trajectory. To achieve fast convergence, a compact expression of the trajectory candidates is needed to reduce the dimension of the search space for PSO. 

\begin{figure}[h] 
	\centering 
	\begin{subfigure}[b]{.5\linewidth} 
		\centering
		\includegraphics[width=1.0\linewidth]{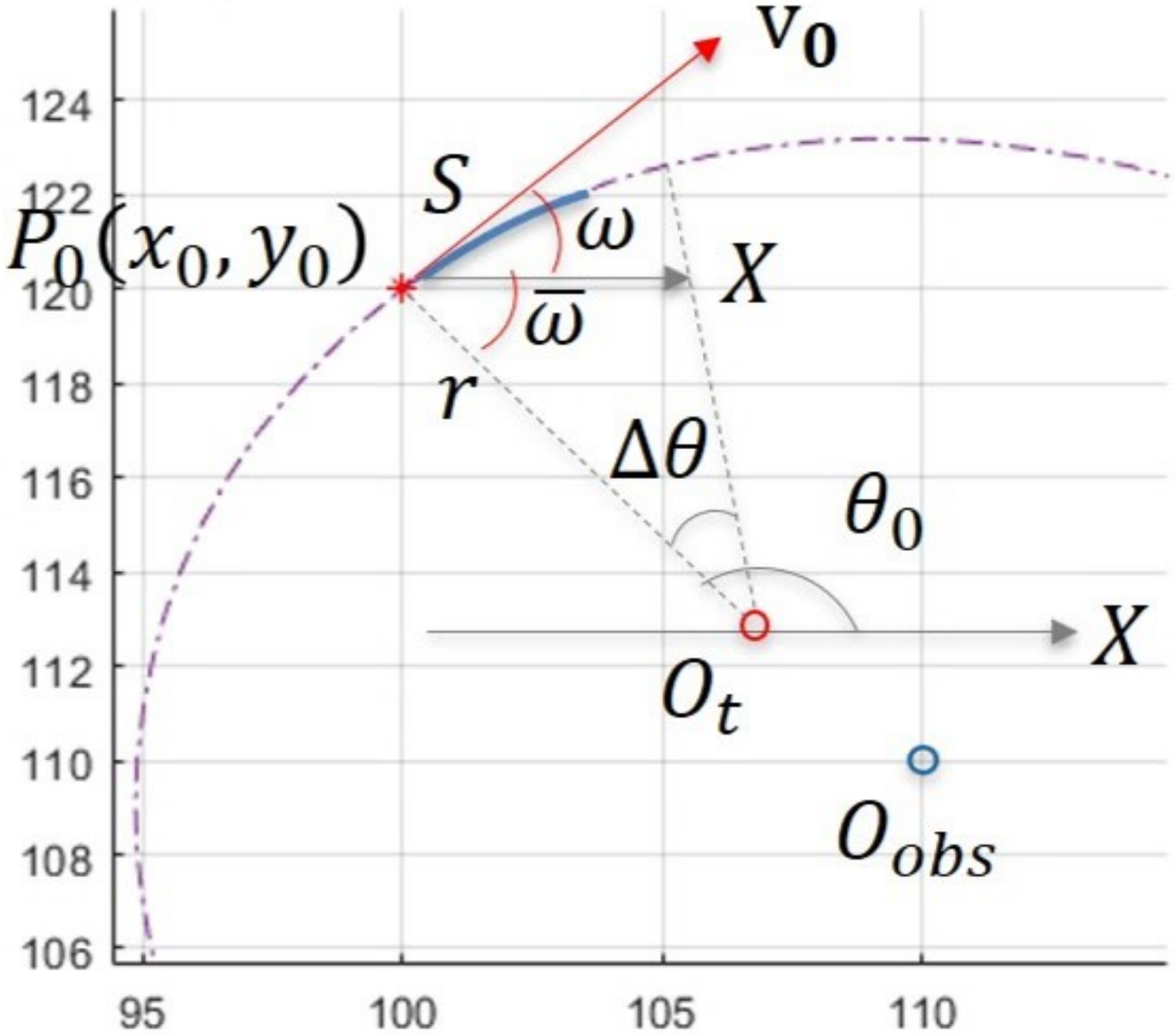} 
		\caption{Parameterization of trajectory. } 
		\label{12a} 
	\end{subfigure}
	\begin{subfigure}[b]{.45\linewidth} 
		\centering
		\includegraphics[width=1.0\linewidth]{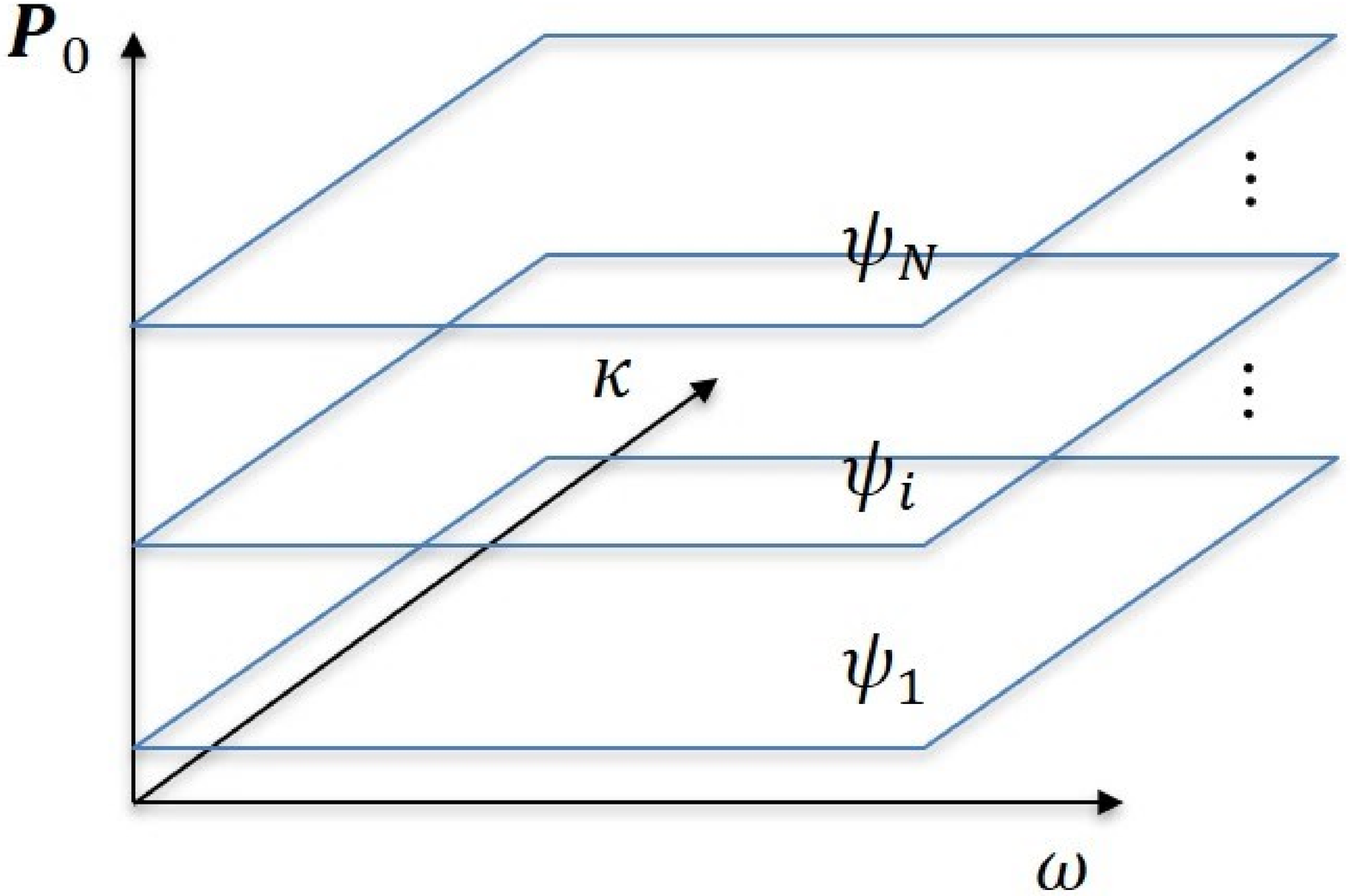} 
		\caption{Search space for PSO. } 
		\label{12b} 
	\end{subfigure} 
	\caption{Parameterization and search space. } 
	\label{fig12} 
\end{figure}

Since each step is small, we use an arc to represent a trajectory as it is more natural for turnings of multi-rotor copters and matches their aerodynamics. 
As illustrated in Fig. \ref{12a}, an arc can be expressed as $S(\omega,\kappa)$ with two parameters where $\kappa=d\omega/dS$ is curvature of the arc and $\omega$ is the slope of the arc relative to the $X$-axis of the coordinate system. 
When $\kappa=0$, it is a straight line along the UAV's current velocity, which means the UAV has no turning. Any point $P_i=[x_i(\omega, \kappa), y_i(\omega, \kappa)]$ on the arc can be expressed by 
\begin{equation} \label{eq16}
\begin{aligned}
&x_i(\omega, \kappa)=\dfrac{\cos(\theta_i)}{\kappa}+\dfrac{\cos(\bar{\omega})}{\kappa}+x_0, \\ 
&y_i(\omega, \kappa)=\dfrac{\sin(\theta_i)}{\kappa}+\dfrac{\sin(\bar{\omega})}{\kappa}+y_0, \\ 
&\theta_i\in[\theta_0-\triangle\theta, \theta_0].
\end{aligned}
\end{equation}
where $\theta_i$ is the slope of the vector from the turning center $O_t$ to a point on the arc, and $\Delta\theta$ is the range of $\theta_i$. 

It can be seen that an arc is uniquely located based on its $[\omega, \kappa]$ and initial point $[x_0,y_0]$. Hence,  a particle can be expressed as $\boldsymbol{\xi}=[\omega, \kappa]_{|[x_0,y_0]}$. Therefore, the search space for trajectory planning is reduced to two dimensional. 
As illustrated in Fig. \ref{12b}, solutions for different UAVs exist in orthogonal sub-spaces where $\psi_i$ is the sub-space for the $i^{th}$ UAV. UAVs in a swarm search for solutions simultaneously.

Based on the representation of trajectories in Eq. (\ref{eq16}), 
the particles can be updated as follows: 
\begin{equation}\label{eq6} 
\begin{aligned} 
&\boldsymbol{v}_i^{t+1}=\mu_0\boldsymbol{v}_i^{t}+\mu_1 (\boldsymbol{p}_i^{t}-\boldsymbol{\xi}_i^{t})+\mu_2 (\boldsymbol{p}_g^{t}-\boldsymbol{\xi}_i^{t}) \\ 
&\boldsymbol{\xi}_i^{t+1}=\boldsymbol{\xi}_i^{t}+\boldsymbol{v}_i^{t+1} 
\end{aligned} 
\end{equation}
where $\boldsymbol{v}_i^t$ and $\boldsymbol{\xi}_i^t$ are the velocity and position of particle $i$ at time $t$, $\boldsymbol{p}_i^{t}$ is the personal best experience of particle $i$ at time $t$ and $\boldsymbol{p}_g^{t}$ is the global best experience of the swarm at time $t$. $\mu_0$ is an inertia weight, $\mu_1$ and $\mu_2$ are acceleration coefficients uniformly distributed in $[0, 1]$ independently.

\section{Performance Evaluation} \label{simul} 

\subsection{Simulation Setup} 

In our simulations, an obstacle is detected when the distance between the obstacle and any UAV is smaller than the detection range of the Lidar sensor $R_d=50m$. A swarm starts to avoid the obstacle when the distance of any UAV to the obstacle is smaller than the avoiding distance $D_a=30m$. 
Since industrial UAVs usually have a maximum ground speed of 10 $m/s$ to 17 $m/s$ \cite{DJIIndust} 
and consumer level UAVs commonly have a ground speed of 5 $m/s$ to 14 $m/s$ \cite{DJIConsum}
, we set $v_s$=10 $m/s$ in all simulations. An obstacle can be either static or dynamic with velocity $v_{obs}$ varying from 0 to 10 $m/s$. In our simulations, the swarm members are  equally spaced in a circle with a radius $\tau=20$ $m$, and the number of UAVs in the swarm $N$ is varied from 2 to 10. 

We compare $E^2Coop$ with the following two schemes:
\begin{itemize}
	\item \textit{FFPSO} \cite{hyb5}: A new term repelling particles from one another is introduced to the velocity update equation in PSO. 
	The fitness function of a particle is simply the negative of the distance between the particle and its destination. 
	\item \textit{PPSO} \cite{hyb6}:  A smoothing field in APF is introduced in order to smooth the trajectories of UAVs. Then, PSO is adopted to find the optimal trajectories on APF. The fitness value of particles is just the field intensity. 
\end{itemize} 

All the three schemes ($E^2Coop$, FFPSO and PPSO) are tested in two scenarios: 1) \textit{Obstacle in Front} where the initial distance between the swarm and the obstacle is 200 $m$, and 2) \textit{Obstacle in Side} where the obstacle appears either on the left-front or right-front side of the swarm, the initial distance between the swarm and the obstacle is 200 $m$, and $v_{obs}$ is perpendicular to $v_{s}$. In our simulations, we observed that the minimum distance between obstacles and trajectories generated by FFPSO and PPSO is 26 m and 16 m, respectively, and the minimum UAV-to-UAV distance in the trajectories generated by FFPSO and PPSO is around 0.5 m. For a fair comparison on energy consumption, we set $D_{obs}$ in $E^2Coop$ to 26 $m$ and 16 $m$, and $D_{v2v}$ in $E^2Coop$ is set to 0.5 $m$.  Each combination is repeated 10 times. 

\subsection{Simulation Results} 

\begin{figure}[t] 
	\centering 
	\begin{subfigure}[b]{.5\linewidth} 
		\centering
		\includegraphics[width=1.0\linewidth]{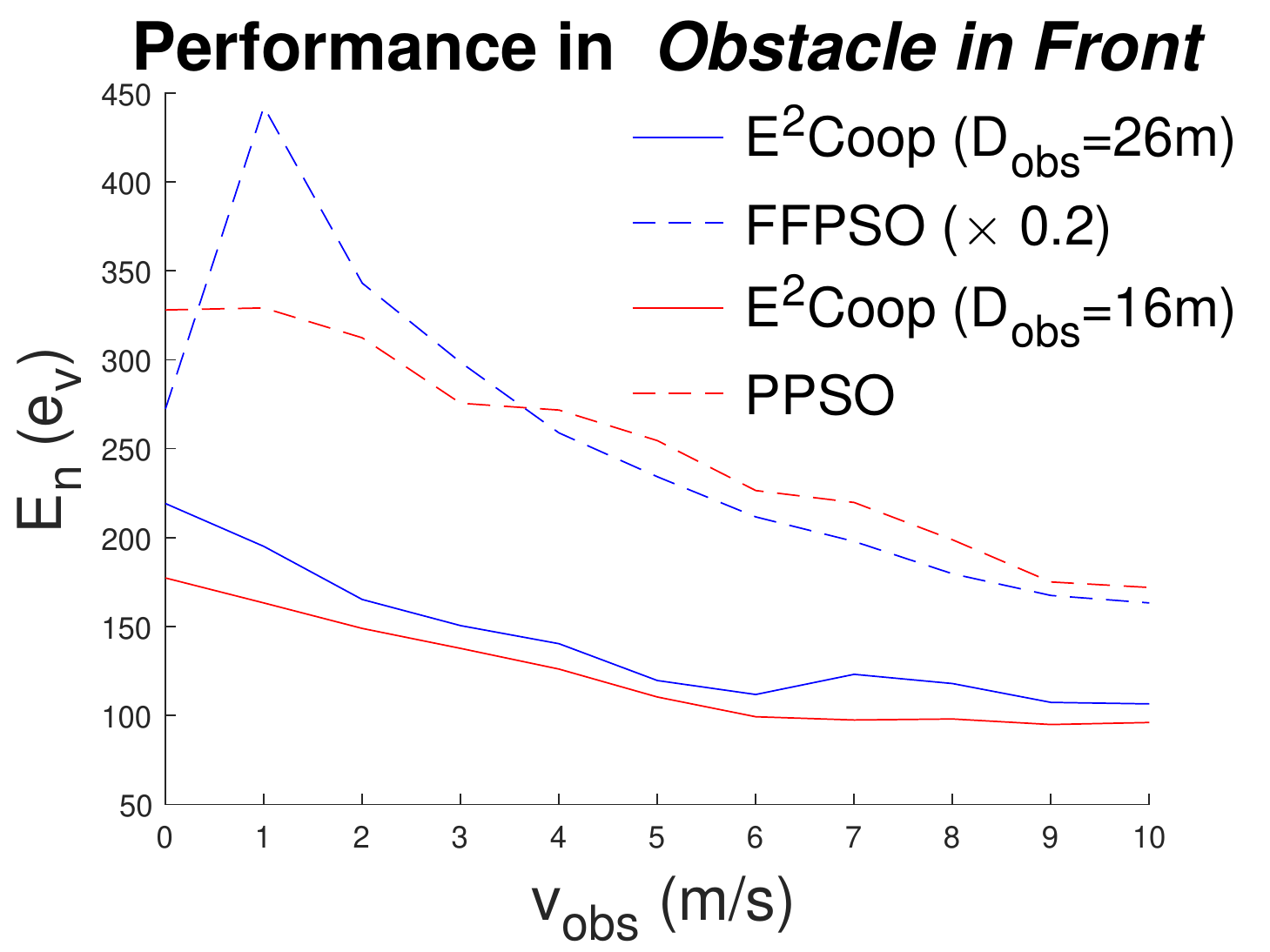} 
		\caption{Obstacle in Front} 
		\label{15a} 
	\end{subfigure}
	\begin{subfigure}[b]{.5\linewidth} 
		\centering
		\includegraphics[width=1.0\linewidth]{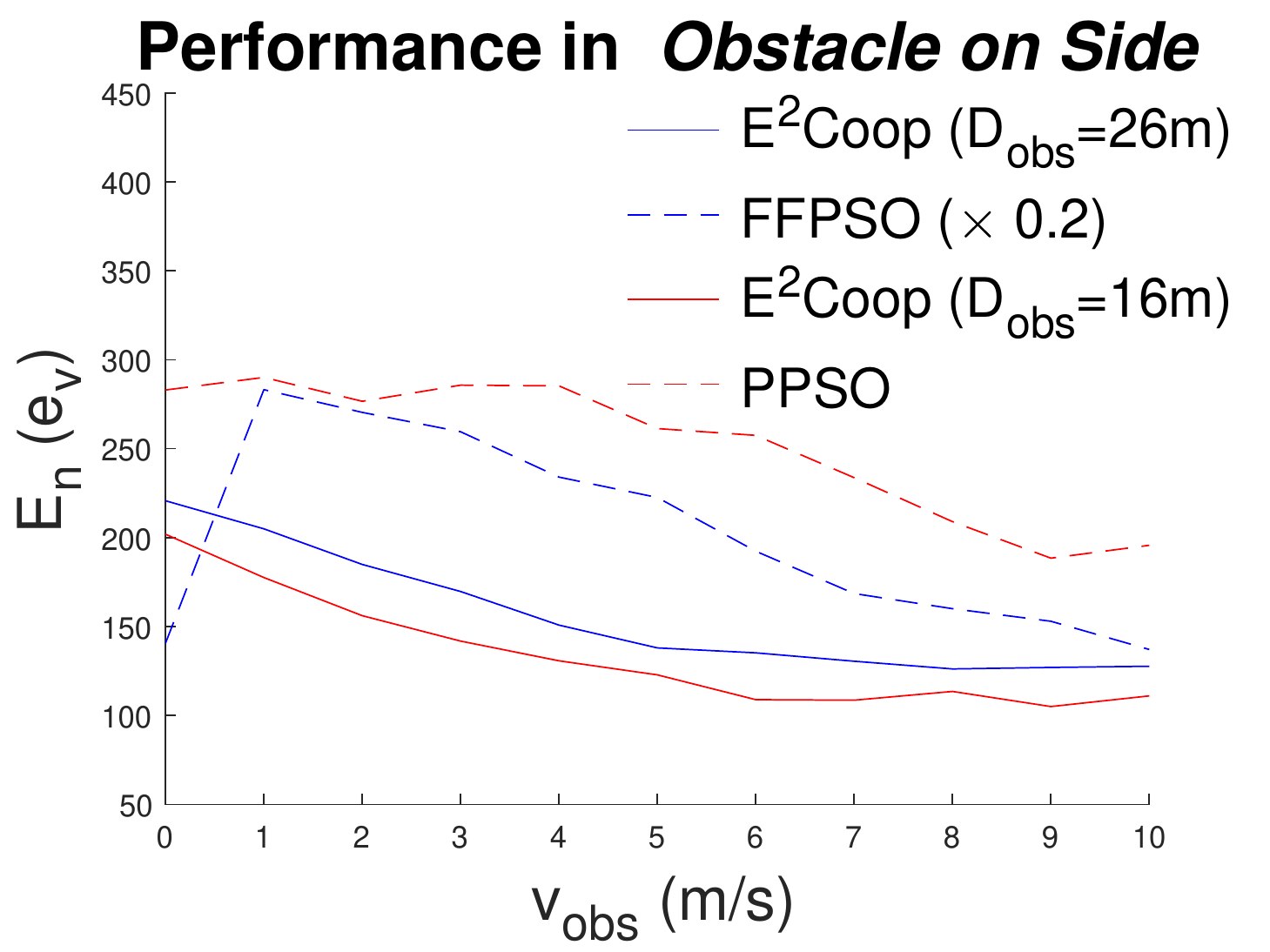} 
		\caption{Obstacle on Side} 
		\label{15b} 
	\end{subfigure} 
	\caption{Energy consumption. } 
	\label{fig16} 
\end{figure} 

Fig.~\ref{fig16} shows the energy consumption for three schemes in both scenarios with the variation of obstacle speed. The energy consumption of FFPSO is scaled by 0.2 for a clear comparison. 
It can be seen that $E^2Coop$ has the lowest energy consumption among all schemes in both scenarios. This is because the fitness function we designed aims to minimize energy consumption by minimizing trajectory's curvature. 
Moreover, the environment field constructed by the swarm provides a global coordination among swarm members for collision avoidance. Hence, the trajectories for individual UAVs do not conflict with each other. FFPSO performs the poorest because there is no such a scheme that smooths the trajectories. 
Although PPSO has a smoothing field in its APF method, each UAV plans its own trajectory without a global coordination. As a result, the trajectories generated by PPSO are still very zigzag. 

For all three schemes, energy consumption decreases as the obstacle's speed increases. This is because that, when the obstacle is at a high speed, the avoidance of UAVs starts early and finishes quickly. This causes their trajectories to be short and have less turnings. An abrupt change can be spot on the curve of FFPSO in both scenarios at $v_{obs}=1~m/s$. This is because the repelling force of obstacles pushes the particles to the direction of $v_{obs}$, and hence causes a very sharp turning on their trajectories. So the energy consumption of FFPSO drastically increases when $v_{obs}$ changes to $1~m/s$. Comparing Fig.~\ref{15a} and Fig. \ref{15b}, the energy consumption of FFPSO is smaller when the obstacle comes from side of the swarm, whereas the other two schemes remain roughly the same. 
This is because an obstacle coming from a side of the swarm pushes the trajectories slightly deviate from $v_s$. 
Overall, the results show that $E^2Coop$ can save up to 51\% energy in \textit{Obstacle in Front} and 46\% in \textit{Obstacle on Side} in comparison with the other two schemes. 

\subsection{Parameter Analysis} 
\begin{figure}[t] 
	\centering 
	\begin{subfigure}[b]{.47\linewidth} 
		\centering
		\includegraphics[width=1.0\linewidth]{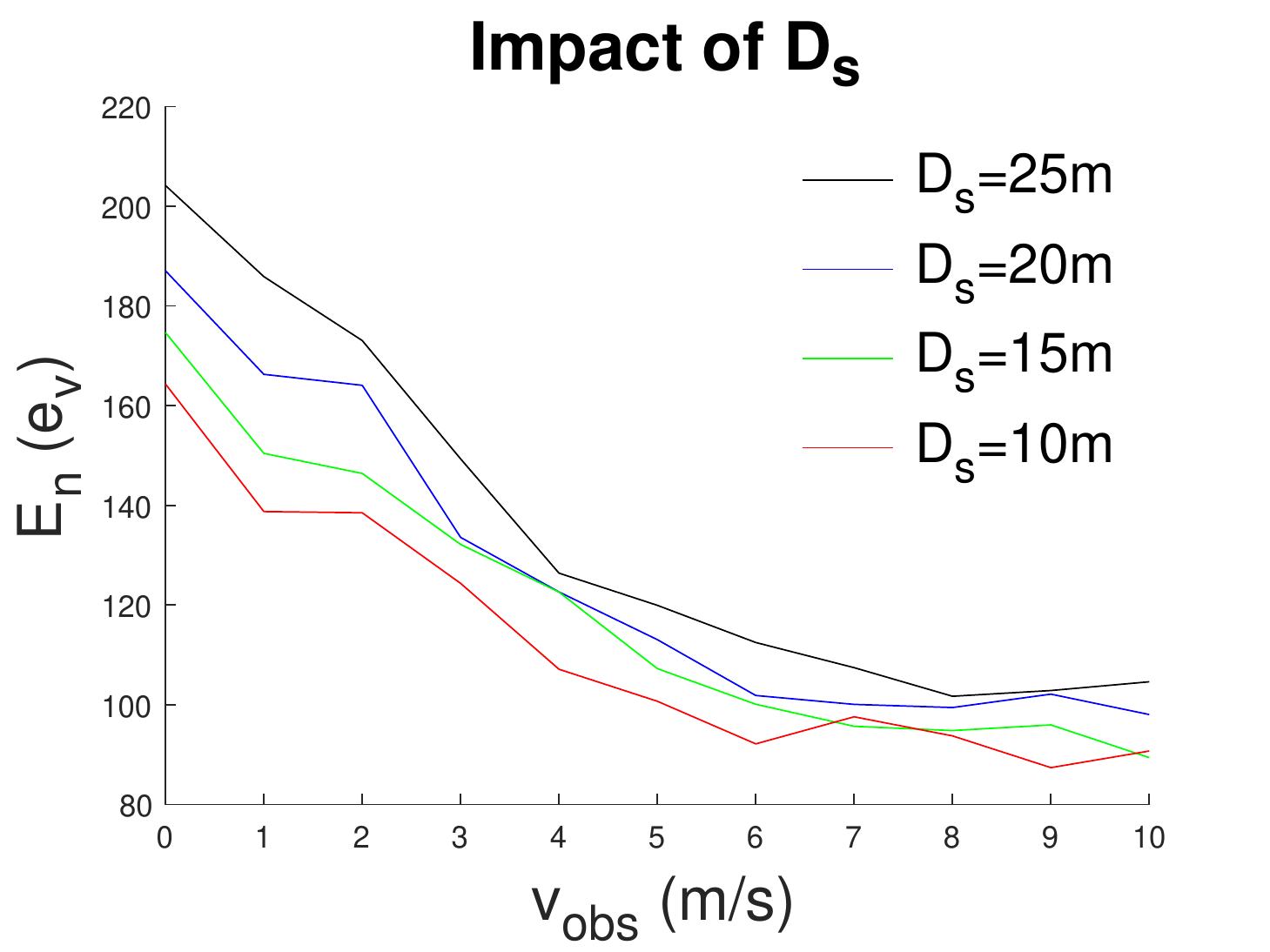} 
		\caption{Impact of $D_{obs}$. } 
		\label{16a} 
	\end{subfigure} 
	\begin{subfigure}[b]{.47\linewidth} 
		\centering
		\includegraphics[width=1.0\linewidth]{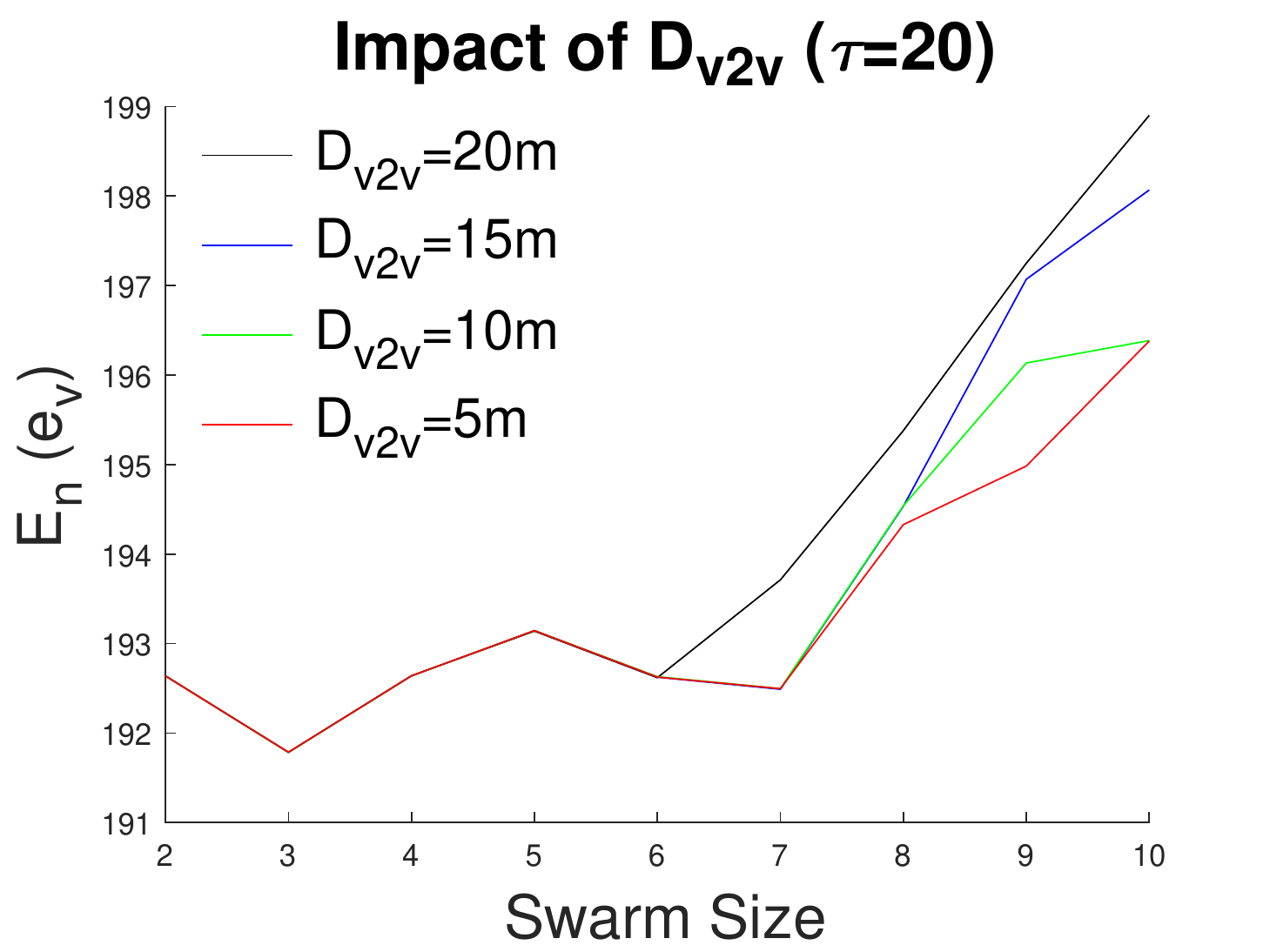} 
		\caption{Impact of $D_{v2v}$. }  
		\label{16b} 
	\end{subfigure} 
	\begin{subfigure}[b]{.47\linewidth} 
		\centering
		\includegraphics[width=1.0\linewidth]{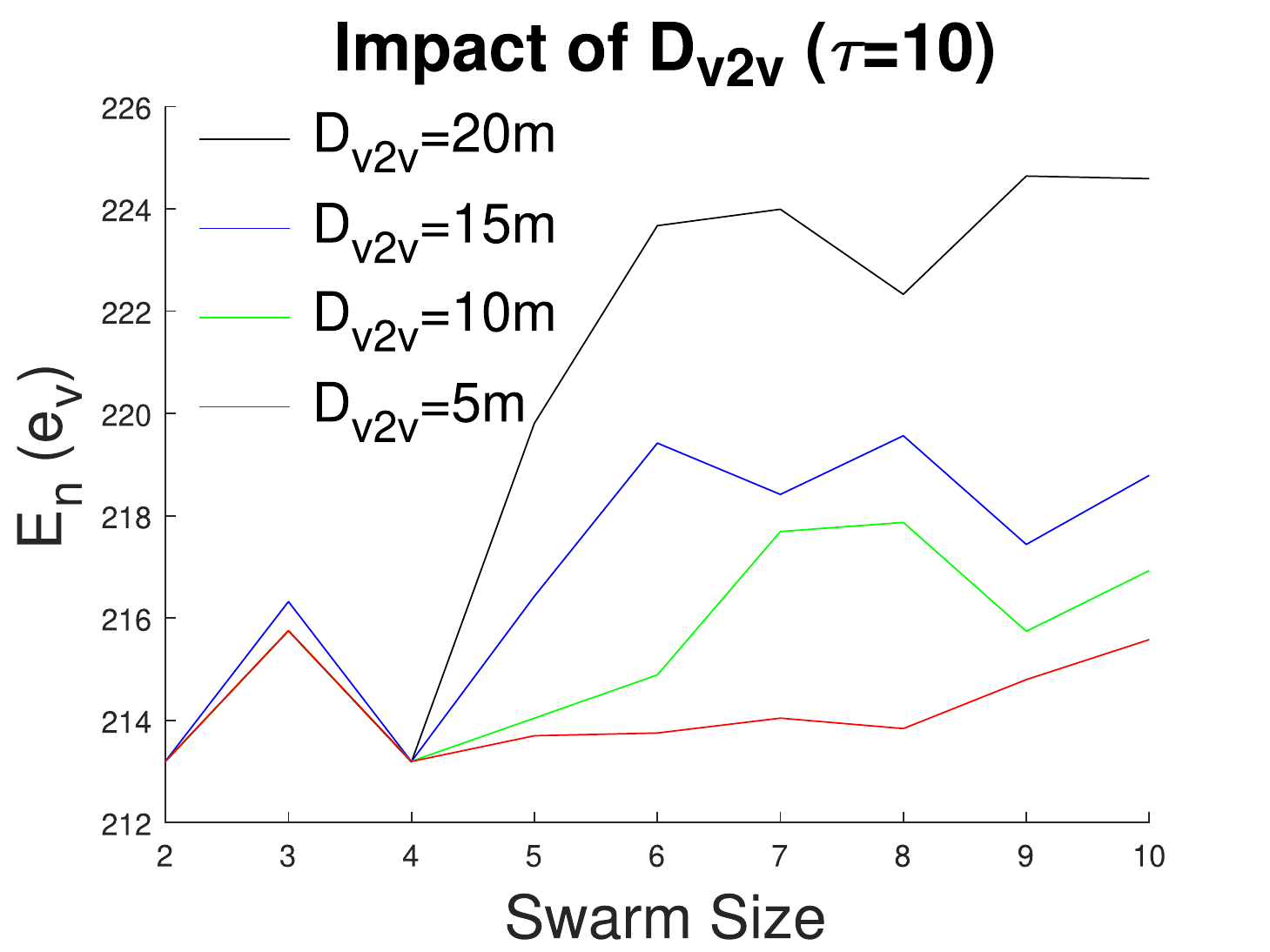} 
		\caption{Impact of $D_{v2v}$. }  
		\label{16c} 
	\end{subfigure} 
	\begin{subfigure}[b]{.47\linewidth} 
		\centering
		\includegraphics[width=1.0\linewidth]{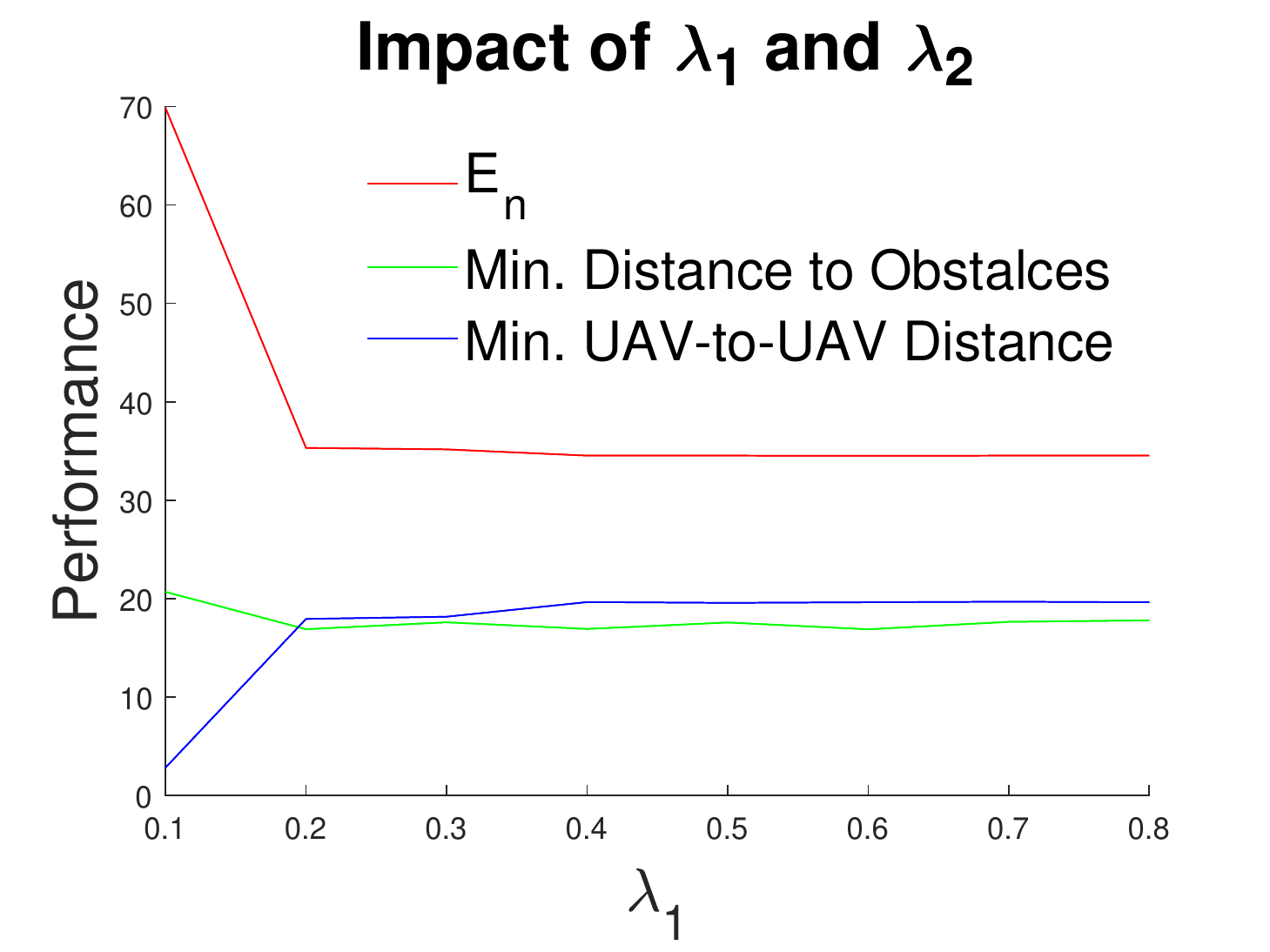} 
		\caption{Impact of $\lambda_1$ and $\lambda_2$. } 
		\label{16d} 
	\end{subfigure} 
	\caption{Parameter analysis results. } 
	\label{fig17} 
\end{figure}

Important parameters of $E^2Coop$ include $D_{obs}, D_{v2v}, \lambda_1$ and $\lambda_2$. 
Given a setting on $v_s$ and $v_{obs}$, large $D_{obs}$ is expected to prolong trajectories, and large $D_{v2v}$ is also expected to prolong and curve trajectories. $\lambda_1$ and $\lambda_2$ are linear weights of energy efficiency and safety concerns in our fitness function \eqref{eq11}. With $\lambda_1+\lambda_2=1$, energy efficiency and safety appear to be a pair of trade-off. We now study the impact of these parameters on the performance of $E^2Coop$ using the scenario of \textit{Obstacle in Front}. Each combination is repeated 3 times. The simulation results are shown in Fig.~\ref{fig17}. 

\subsubsection{Impact of $D_{obs}$} 
From Fig.~\ref{16a}, it can be seen that energy consumption of $E^2Coop$ increases with the increase of $D_{obs}$ under all obstacle speeds. This is because larger $D_{obs}$ forces the trajectories to go further away from the obstacle. Therefore, the curvature and length of the trajectories are larger, and hence more energy consumption is required. 

\subsubsection{Impact of $D_{v2v}$} 
From Fig.~\ref{16b} and Fig.~\ref{16c}, it can be seen that energy consumption of $E^2Coop$ increases with swarm size. This is because UAV members must remain a certain distance away from each other during avoidance. Trajectories of individual UAVs have to be longer and more curved when more members are in a swarm. 
Fig.~\ref{16b} and Fig.~\ref{16c} show the energy consumption when the swarms are in a formation of circle with a radius of 20 and 10 $m$, respectively. In Fig.~\ref{16b}, energy consumption of $E^2Coop$ starts to increase with $D_{v2v}$ from $N=6$. While in Fig.~\ref{16b}, energy consumption of $E^2Coop$ starts to increase with $D_{v2v}$ from $N=4$. This is because the impact of $D_{v2v}$ on energy is small when swarm size is small. The impact of $D_{v2v}$ on energy is large when UAVs are close to each other. 

\subsubsection{Impact of $\lambda_1$ and $\lambda_2$} 
From Fig.~\ref{16d}, we observe that in extreme cases when $\lambda_1\leq 0.1$, $E^2Coop$ is still able to find trajectories to the destination. However, the trajectories have many turnings as the smoothing term in our fitness function has a small weight. Hence, energy consumption is high. What's more, the minimum UAV-to-UAV distance is very small, while the minimum distance to obstacles is relatively large. This implies that in such extreme cases, UAVs can avoid obstacles, but UAV members may have the risk to collide with each other. 
When $\lambda_1$ is larger than 0.4, energy consumption, minimum distance to obstacles, and minimum UAV-to-UAV distance become stable and converge to a certain level. 
When $\lambda_1=0.9$ ($\lambda_2=0.1$), we observe that $E^2Coop$ fails to find trajectories to destination. This is because the second term in our fitness function has a small weight, so UAVs can't find contours on the environment field. It can be seen that the performance of $E^2Coop$ is not very sensitive to the selection of $\lambda_1$ and $\lambda_2$. 

\subsection{ODA Demonstration}
\begin{figure}[t] 
	\centering
	\begin{subfigure}[b]{.47\linewidth}
		\centering
		\includegraphics[width=1.0\linewidth]{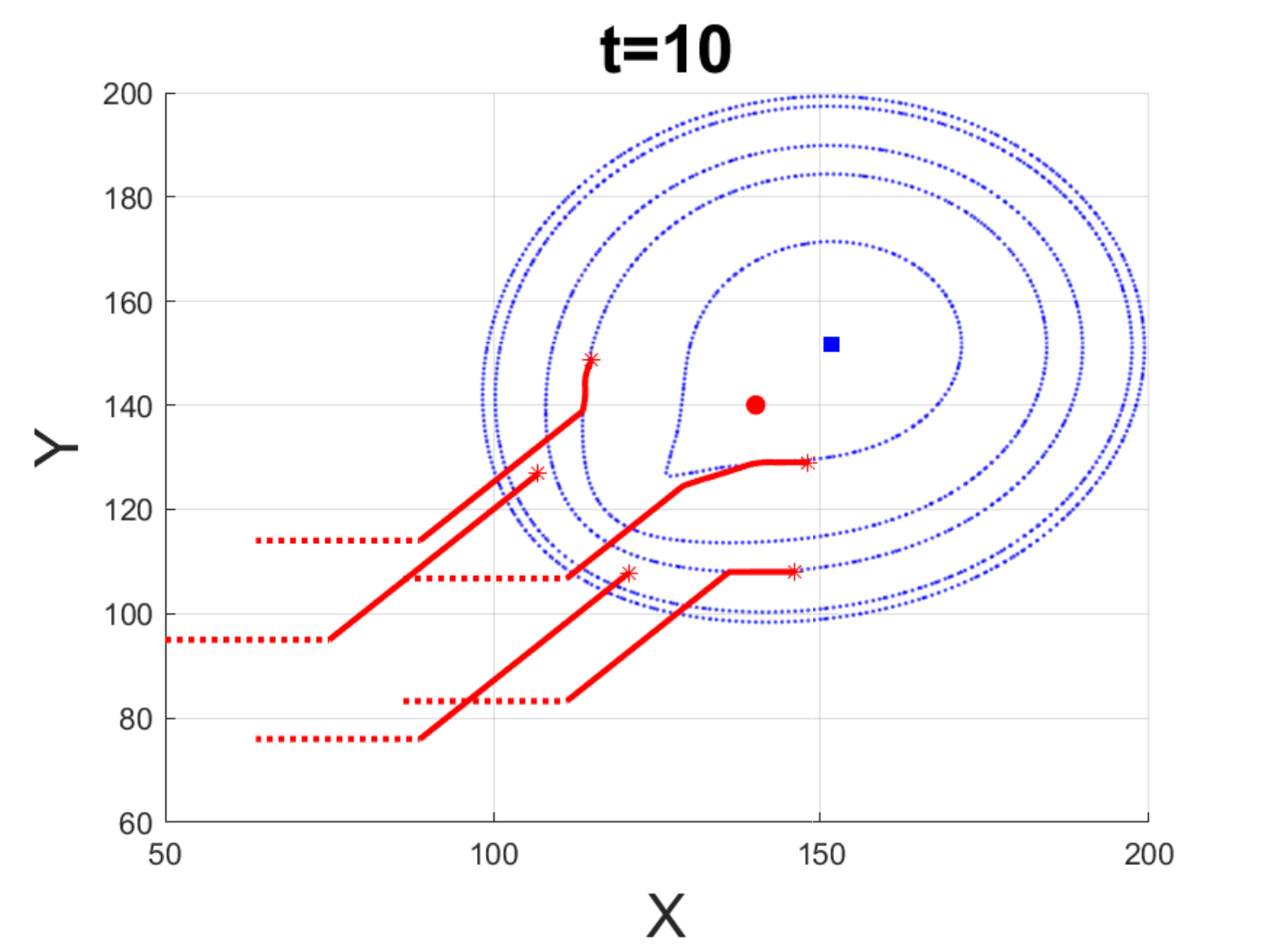} 
		\caption{Start avoidance. } 
		\label{11b} 
	\end{subfigure} 
	\begin{subfigure}[b]{0.47\linewidth}
		\centering
		\includegraphics[width=1.0\linewidth]{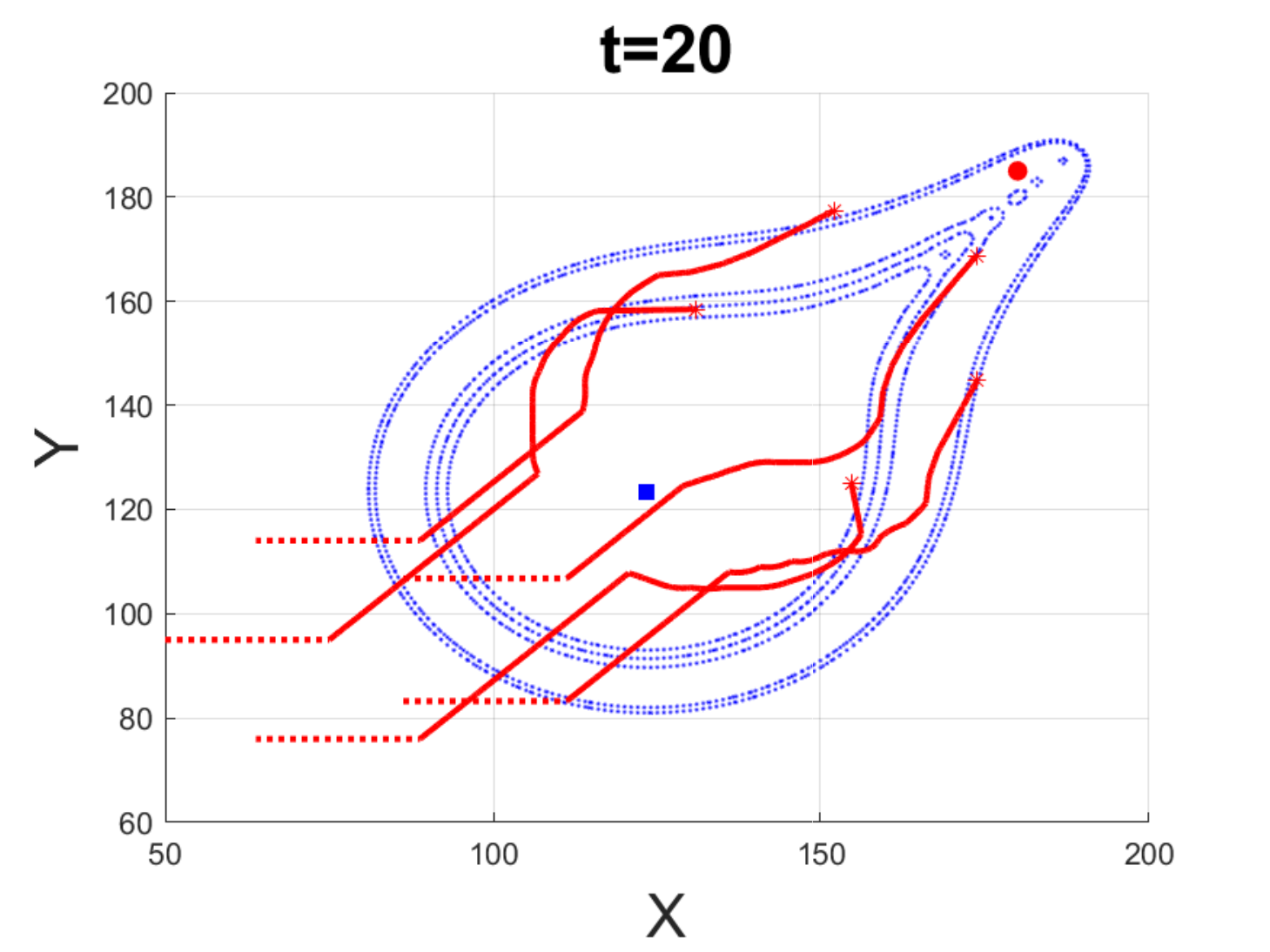} 
		\caption{After avoidance. } 
		\label{11d} 
	\end{subfigure} 
	\begin{subfigure}[b]{0.47\linewidth}
		\centering
		\includegraphics[width=1.0\linewidth]{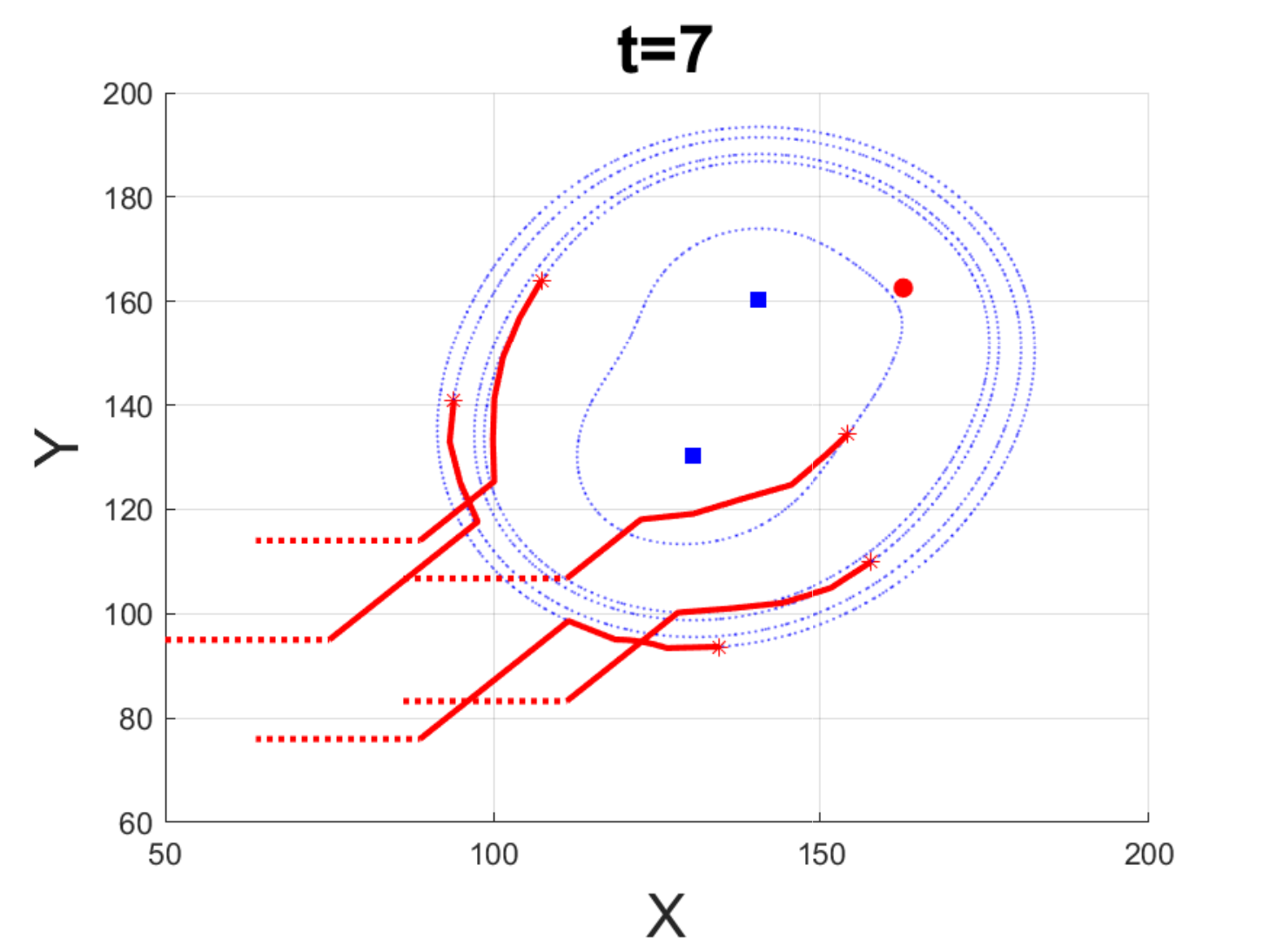} 
		\caption{Start avoidance. } 
		\label{11f} 
	\end{subfigure} 
	\begin{subfigure}[b]{0.47\linewidth}
		\centering
		\includegraphics[width=1.0\linewidth]{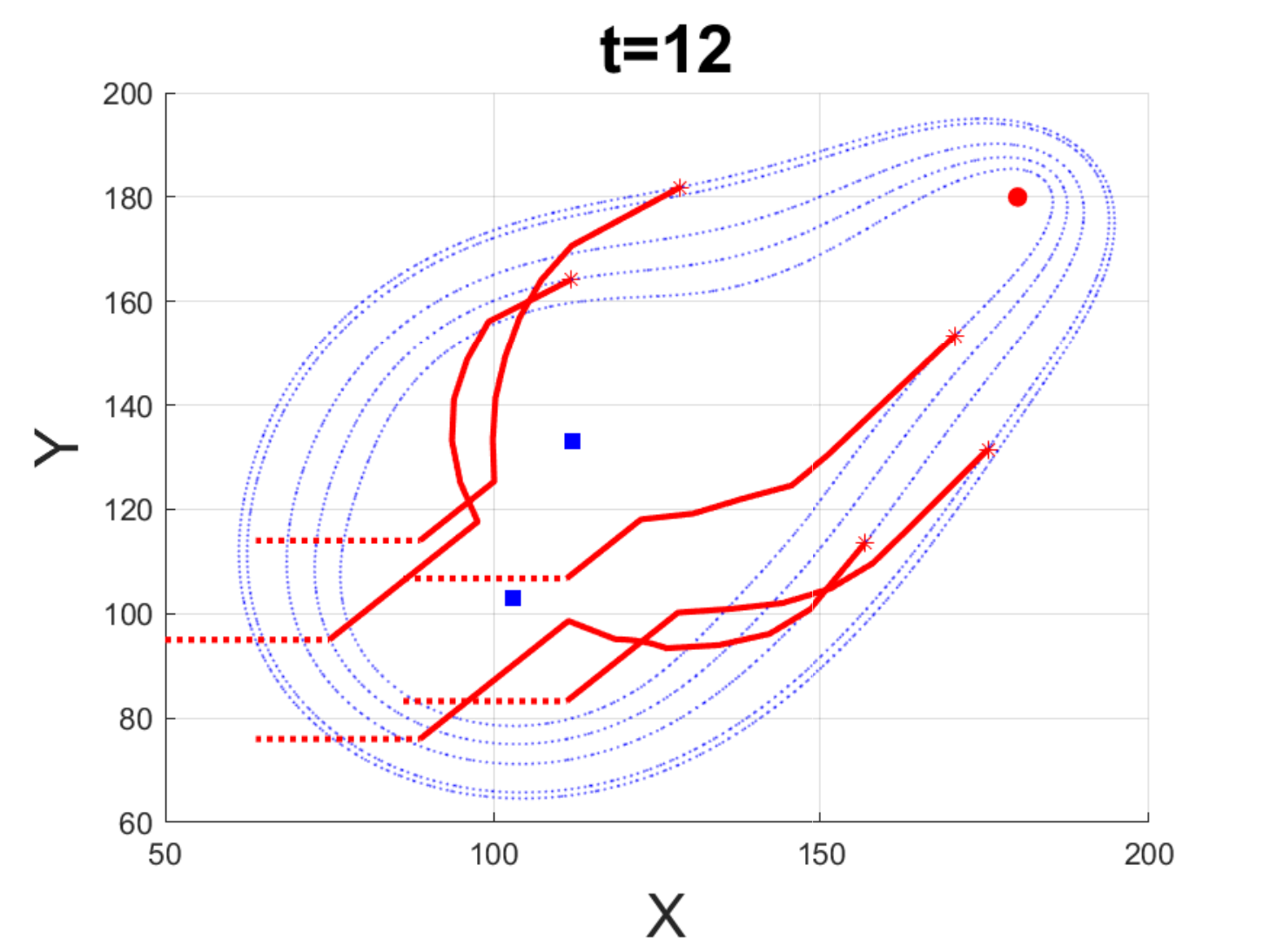} 
		\caption{After avoidance. } 
		\label{11h} 
	\end{subfigure} 
	\caption{Demonstration of generated trajectories. }
	\label{fig11} 
\end{figure}

Fig.~\ref{fig11} shows the trajectories generated by $E^2Coop$, where $\lambda_1=\lambda_2=0.5$. Five UAVs are in a swarm with $v_s=5m/s$ and $v_{obs}=10m/s$. $D_{v2v}$ and $D_{obs}$ are set to 10$m$ and 5$m$, respectively. Fig. \ref{11b} and \ref{11d} are single-obstacle scenario, while Fig. \ref{11f} and \ref{11h} are double-obstacle scenario. 
The virtual leader of the swarm and geometrical center of the obstacle are shown in red circle and blue square, respectively. 
Planned trajectories are shown in red solid curves. Contours are shown in blue solid curves to help reflect the deformation of the environment field. 

\section{Conclusions} \label{concl} 
In this paper we investigate the problem of trajectory planning for UAV swarms to avoid collisions in an energy efficient manner. $E^2Coop$, a hybrid scheme that combines APF and PSO is proposed. $E^2Coop$ is designed to take advantage of the search ability of PSO and the environmental representations of APF to coordinate the trajectory planning of swarm members in a global perspective. Results demonstrate that $E^2Coop$ can generate safe trajectories and consume much less energy than the compared schemes. Regarding large-scale aerial applications, future work can be done to extend it for three dimensional trajectory planning.  

\section{ Acknowledgments}
The authors would like to thank the Air Traffic Management Research Institute, Nanyang Technological University, Singapore for helping conduct the field trials. 

\fontsize{9.8pt}{10.8pt}\selectfont 
\bibliography{references}

\begin{thebibliography}{21}
\providecommand{\natexlab}[1]{#1}
\providecommand{\url}[1]{\texttt{#1}}
\providecommand{\urlprefix}{URL }
\expandafter\ifx\csname urlstyle\endcsname\relax
  \providecommand{\doi}[1]{doi:\discretionary{}{}{}#1}\else
  \providecommand{\doi}{doi:\discretionary{}{}{}\begingroup
  \urlstyle{rm}\Url}\fi

\bibitem[{ATMRI(2020)}]{fieldTrials}
ATMRI. 2020.
\newblock Field trials full video.
\newblock \url{https:/https://youtu.be/EyZzIUlmAlI}.
\newblock (Accessed on 03/05/2021).

\bibitem[{Cai and Yang(2014)}]{hyb6}
Cai, Y.; and Yang, S.~X. 2014.
\newblock A potential field-based PSO approach for cooperative target searching
  of multi-robots.
\newblock In \emph{Proceeding of the 11th World Congress on Intelligent Control
  and Automation}, 1029--1034. IEEE.

\bibitem[{DJI(2020{\natexlab{a}})}]{DJIIndust}
DJI. 2020{\natexlab{a}}.
\newblock MATRICE 300 RTK - Specifications - DJI.
\newblock \url{https://www.dji.com/nz/matrice-300/specs}.
\newblock (Accessed on 03/17/2021).

\bibitem[{DJI(2020{\natexlab{b}})}]{DJIConsum}
DJI. 2020{\natexlab{b}}.
\newblock Mavic Air 2 - Specifications - DJI.
\newblock \url{https://www.dji.com/nz/mavic-air-2/specs}.
\newblock (Accessed on 03/17/2021).

\bibitem[{Huang and Low(2018)}]{b11}
Huang, S.; and Low, K. 2018.
\newblock A Path Planning Algorithm for Smooth Trajectories of Unmanned Aerial
  Vehicles via Potential Fields.
\newblock In \emph{2018 15th International Conference on Control, Automation,
  Robotics and Vision (ICARCV)}, 1677--1684. IEEE.

\bibitem[{Kass, Witkin, and Terzopoulos(1988)}]{b5}
Kass, M.; Witkin, A.; and Terzopoulos, D. 1988.
\newblock Snakes: Active contour models.
\newblock \emph{International journal of computer vision} 1(4): 321--331.

\bibitem[{Liang et~al.(2019)Liang, Qu, Wang, Li, and Zhang}]{b37}
Liang, X.; Qu, X.; Wang, N.; Li, Y.; and Zhang, R. 2019.
\newblock Swarm control with collision avoidance for multiple underactuated
  surface vehicles.
\newblock \emph{Ocean Engineering} 191: 106516.

\bibitem[{Liu et~al.(2018)Liu, Jiang, Xu, Cheng, Xie, and Lin}]{b25}
Liu, Z.; Jiang, Z.; Xu, T.; Cheng, H.; Xie, Z.; and Lin, L. 2018.
\newblock Avoidance of High-Speed Obstacles Based on Velocity Obstacles.
\newblock In \emph{2018 IEEE International Conference on Robotics and
  Automation (ICRA)}, 7624--7630. IEEE.

\bibitem[{Parker, Butterworth, and Luo(2019)}]{hyb5}
Parker, L.; Butterworth, J.; and Luo, S. 2019.
\newblock Fly safe: Aerial swarm robotics using force field particle swarm
  optimisation.
\newblock \emph{arXiv preprint arXiv:1907.07647} .

\bibitem[{Pradhan et~al.(2019)Pradhan, Nandi, Hui, Roy, and Rodrigues}]{hyb8}
Pradhan, B.; Nandi, A.; Hui, N.~B.; Roy, D.~S.; and Rodrigues, J. 2019.
\newblock A Novel Hybrid Neural Network Based Multi Robot Path Planning with
  Motion Coordination.
\newblock \emph{IEEE Transactions on Vehicular Technology} .

\bibitem[{Shan et~al.(2020)Shan, Luo, Xiong, Wu, and Li}]{energy}
Shan, F.; Luo, J.; Xiong, R.; Wu, W.; and Li, J. 2020.
\newblock Looking before Crossing: An Optimal Algorithm to Minimize UAV Energy
  by Speed Scheduling with a Practical Flight Energy Model.
\newblock In \emph{IEEE INFOCOM 2020-IEEE Conference on Computer
  Communications}, 1758--1767. IEEE.

\bibitem[{SICK(2020)}]{lidar}
SICK. 2020.
\newblock TiM561-2050101 | Detection and ranging solutions | SICK.
\newblock
  \url{https://www.sick.com/us/en/detection-and-ranging-solutions/2d-lidar-sensors/tim5xx/tim561-2050101/p/p369446}.
\newblock (Accessed on 03/17/2021).

\bibitem[{Snape et~al.(2009)Snape, Van Den~Berg, Guy, and Manocha}]{vo2}
Snape, J.; Van Den~Berg, J.; Guy, S.~J.; and Manocha, D. 2009.
\newblock Independent navigation of multiple mobile robots with hybrid
  reciprocal velocity obstacles.
\newblock In \emph{2009 IEEE/RSJ International Conference on Intelligent Robots
  and Systems}, 5917--5922. IEEE.

\bibitem[{Snape et~al.(2011)Snape, Van Den~Berg, Guy, and Manocha}]{b23}
Snape, J.; Van Den~Berg, J.; Guy, S.~J.; and Manocha, D. 2011.
\newblock The hybrid reciprocal velocity obstacle.
\newblock \emph{IEEE Transactions on Robotics} 27(4): 696--706.

\bibitem[{Song et~al.(2019)Song, Wang, Zou, Xu, and Alsaadi}]{pso4}
Song, B.; Wang, Z.; Zou, L.; Xu, L.; and Alsaadi, F.~E. 2019.
\newblock A new approach to smooth global path planning of mobile robots with
  kinematic constraints.
\newblock \emph{International Journal of Machine Learning and Cybernetics}
  10(1): 107--119.

\bibitem[{Stolaroff et~al.(2018)Stolaroff, Samaras, O’Neill, Lubers,
  Mitchell, and Ceperley}]{b28}
Stolaroff, J.~K.; Samaras, C.; O’Neill, E.~R.; Lubers, A.; Mitchell, A.~S.;
  and Ceperley, D. 2018.
\newblock Energy use and life cycle greenhouse gas emissions of drones for
  commercial package delivery.
\newblock \emph{Nature communications} 9(1): 409.

\bibitem[{Tan et~al.(2019)Tan, Wang, Ong, and Low}]{algo1}
Tan, Q.; Wang, Z.; Ong, Y.-S.; and Low, K.~H. 2019.
\newblock Evolutionary optimization-based mission planning for UAS traffic
  management (UTM).
\newblock In \emph{2019 International Conference on Unmanned Aircraft Systems
  (ICUAS)}, 952--958. IEEE.

\bibitem[{Tharwat et~al.(2019)Tharwat, Elhoseny, Hassanien, Gabel, and
  Kumar}]{pso3}
Tharwat, A.; Elhoseny, M.; Hassanien, A.~E.; Gabel, T.; and Kumar, A. 2019.
\newblock Intelligent B{\'e}zier curve-based path planning model using Chaotic
  Particle Swarm Optimization algorithm.
\newblock \emph{Cluster Computing} 22(2): 4745--4766.

\bibitem[{Tsang et~al.(2018)Tsang, Ni, Wong, and Shi}]{b12}
Tsang, K. F.~E.; Ni, Y.; Wong, C. F.~R.; and Shi, L. 2018.
\newblock A Novel Warehouse Multi-Robot Automation System with Semi-Complete
  and Computationally Efficient Path Planning and Adaptive Genetic Task
  Allocation Algorithms.
\newblock In \emph{2018 15th International Conference on Control, Automation,
  Robotics and Vision (ICARCV)}, 1671--1676. IEEE.

\bibitem[{Wallar and Plaku(2013)}]{reviewer1}
Wallar, A.; and Plaku, E. 2013.
\newblock Path planning for swarms by combining probabilistic roadmaps and
  potential fields.
\newblock In \emph{Conference Towards Autonomous Robotic Systems}, 417--428.
  Springer.

\bibitem[{Xia et~al.(2020)Xia, Han, Zhao, and Wang}]{hyb9}
Xia, G.; Han, Z.; Zhao, B.; and Wang, X. 2020.
\newblock Local Path Planning for Unmanned Surface Vehicle Collision Avoidance
  Based on Modified Quantum Particle Swarm Optimization.
\newblock \emph{Complexity} 2020.

\end{thebibliography}

\end{document}